\documentclass[11pt]{article}

\usepackage[utf8]{inputenc}
\usepackage[T1]{fontenc}
\usepackage{lmodern}
\usepackage{amsmath, amssymb, amsfonts}
\usepackage{graphicx}
\usepackage{booktabs}
\usepackage{hyperref}
\usepackage{natbib}
\usepackage[margin=1in]{geometry}
\usepackage{microtype}
\usepackage{caption}
\usepackage[most]{tcolorbox}
\usepackage{soul}

\tcbset{
  turn/.style={
    boxrule=0.4pt, arc=1.5pt,
    left=8pt, right=8pt, top=5pt, bottom=5pt,
    fontupper=\small,
    before skip=5pt, after skip=2pt,
    toptitle=2pt, bottomtitle=2pt, coltitle=black!65,
  },
}
\newtcolorbox{systurn}{turn, colback=gray!6,  colframe=gray!45,
  colbacktitle=gray!15,  title={\scriptsize\textsc{System}}}
\newtcolorbox{userturn}{turn, colback=blue!5,  colframe=blue!35!gray,
  colbacktitle=blue!12,  title={\scriptsize\textsc{User}}}
\newtcolorbox{asstturn}{turn, colback=green!5, colframe=green!30!gray,
  colbacktitle=green!10, title={\scriptsize\textsc{Assistant}}}

\newcommand{\prefillmark}[1]{{\sethlcolor{yellow!35}\hl{#1}}}

\usepackage{xcolor}
\usepackage{float}
\usepackage{multirow}
\usepackage{rotating}
\usepackage{enumitem}
\usepackage{tikz}
\usepackage{pgfplots}
\pgfplotsset{compat=1.18}
\usetikzlibrary{arrows.meta, positioning, fit, backgrounds, calc, decorations.pathreplacing}

\newcommand{\Pme}{P(\text{Me})}
\newcommand{\Ent}{\mathrm{H}}
\newcommand{\Surp}{\mathrm{S}}

\title{From Simulation to Enaction: Post-trained Language Models Recognize and React to their own Generations}

\author{%
  Asvin G.$^{1}$\thanks{Correspondence to \texttt{gasvinseeker94@gmail.com}}%
  \quad Jack Lindsey$^{2}$ \\[2pt]
  $^{1}$Institute for Advanced Study, Princeton \quad $^{2}$Anthropic
}

\date{}

\begin{document}

\maketitle

\begin{abstract}
Language models are pretrained as passive predictors with no incentive to model the consequences of their own outputs. Post-training changes this: a model producing its own responses can benefit from recognizing that it is on-policy. We present evidence that post-trained models recognize their on-policy generations, and this recognition is implicitly encoded in their output distributions. In particular, on-policy output distribution entropy is 3--4$\times$ lower than off-policy entropy, across model families and size classes. We trace part of this effect to an internal representation of input surprise, tracking the unlikeliness of the most recent input token according to the model's prior predictions, that causally modulates output entropy. One example of these phenomena can be observed in response to open-ended prompts; post-trained models (unlike pretrained models) collapse their uncertainty over the topic of their upcoming response before the first output token; violating this cached intention with a different-topic prefill results in higher output entropy. We also tested whether models can distinguish on-policy contexts from prefills via explicit verbal report. We find that they can, but that interestingly, this explicit recognition routes through a different mechanism than implicit recognition.
\end{abstract}

\begin{center}
\small Code and figure-generation scripts: \url{https://github.com/asving/Learning-to-be-an-agent}
\end{center}

\begin{figure}[H]
\centering
\includegraphics[width=0.95\textwidth]{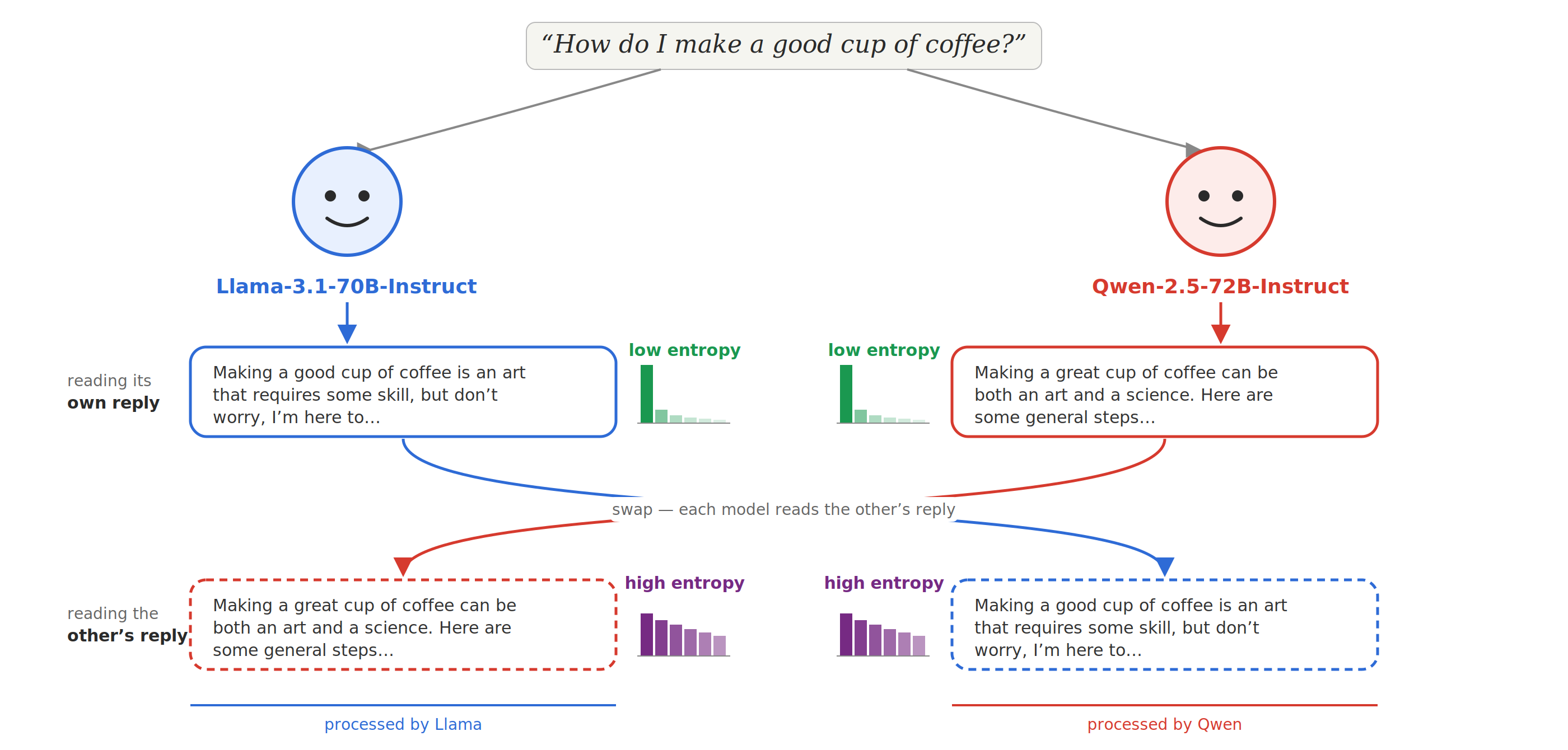}
\caption{Llama and Qwen produce similar-sounding replies to the same question. Yet each model continues its own response with greater confidence (lower entropy) than when prefilled with the other model's response. We investigate this ``self-recognition'' effect in depth in the paper.}
\label{fig:cartoon}
\end{figure}

\section{Introduction}

Language models are initially trained (during pretraining) as next-token predictors. The training aims to minimize the cross-entropy with respect to a fixed data distribution. The irreducible uncertainty of this predictive task bounds how confident the model can be in its predictions at any given time: since many continuations are plausible, the model must spread probability mass across them. Importantly, during pre-training, the model never sees the consequences of its outputs; there is no feedback loop from action to sensory input. The distribution it learns is one it cannot affect, and with no way to influence its future inputs, there is no incentive to model the consequences of its own actions or recognize its own generations. It remains a passive observer of the external distribution.

A natural view of this kind of model, articulated by \citet{janus2022simulators} and developed further by \citet{shanahan2023roleplay, shanahan2023talking}, is that it is a \emph{simulator}: a system that can run any of the characters latent in its training data, with no particular identity of its own. On this view, post-training selects one character---the Assistant---out of many, triggered by user/assistant dialogue formatting, and the model has no special identification with this character~\citep{marks2026psm}.

In principle, however, post-training could reshape the relationship between the model and the Assistant character. Post-training, whether supervised learning or reinforcement learning, typically \emph{only} trains the Assistant's outputs, breaking the symmetry between the Assistant and other characters. In addition, these outputs may be selected on the basis of criteria other than likelihood under a data distribution---for instance, performance on tasks, or relative to human feedback. As the model receives more and more training on how to enact a particular character, in a more goal-oriented fashion than during pretraining, its relationship to that character might change.

We might hypothesize  that a post-trained model transitions from simulation to something more like \emph{enaction}: rather than holding a character at arm's length while making predictions about it, an enacting agent embodies the character, recognizing that its internal states are determinative of future outputs and that those outputs are actions that will influence its own future inputs. We can predict several consequences for a model operating under this paradigm. The model should be able to recognize when it is acting, i.e., when its past trajectory is \emph{on-policy}, and modulate its behavior in response. For instance, when acting, the model might benefit from maintaining more deterministic output distributions, in order to minimize the noise from auto-regressive sampling. We might also expect enacting agents to form more opinionated plans about their future outputs, even when there are multiple reasonable responses they could give. 

In this paper, we provide evidence consistent with these predictions. We find that post-training endows the model with an enhanced ability to distinguish contexts in which it is acting (as the Assistant) from those in which it is passively consuming text. In particular, the model can recognize \emph{implicitly} when its inputs come from its own policy, and it manifests this recognition by modulating its outputs to be significantly lower entropy. Prior work has documented this entropy reduction as a global byproduct of alignment training, framing it as a loss of generation diversity~\citep{kirk2024rlhf_diversity,park2025clip_entropy}; we find instead that the collapse is sharply context-dependent---concentrated in the assistant role and amplified when the model reads its own prior outputs, and strongest under the default Assistant persona relative to other system-prompted characters. We investigate the sources of this effect, finding that it only emerges at sufficiently large model scales, and that off-policy SFT and DPO are sufficient to produce it. We also provide evidence that the model's output entropy is modulated by an internal representation of input surprise, which could (at least in part) account for the self-recognition effect.

In tandem with the token-level collapse in next-token-prediction entropy, we also find that ``\emph{semantic entropy}''---as measured by the spread over the topic the model will discuss in response to an ambiguous prompt---is much lower for the post-trained model than the base model, and that prefills which deviate from the planned topic increase the output entropy, in line with an account where surprise drives entropy increases. We also find that models can \emph{explicitly} report when their response has been prefilled; however, notably, we show that this explicit capability routes through a distinct mechanism which is only invoked on an as-needed basis immediately prior to the prefill detection assessment.

Overall, we read this cluster of findings as a shift from \emph{simulation} toward \emph{enaction}. Under this picture, entropy can be thought of as measuring the cost the model pays for simulation of others besides itself. From this perspective, entropy may also be used as a measure of how strongly the model's self-recognition is entangled with particular personas; empirically, the default Assistant persona appears privileged to an extent. We suspect that more deeply understanding these phenomena and their underlying mechanisms will be important to understanding sophisticated forms of agency and situational awareness in language models.
\section{Entropy as a measure of implicit on-policy recognition}
\label{sec:entropy}

It is well known that post-training reduces the entropy of model generations. Prior work has characterized this as entropy collapse: a loss of diversity that accompanies alignment training~\citep{kirk2024rlhf_diversity, cui2025entropy_mechanism}. The clipping mechanisms in PPO and GRPO drive entropy down even with random rewards~\citep{park2025clip_entropy}. Here, we observed that this entropy reduction is \emph{contingent} on the model being in a specific context. In particular, the chat formatting, and especially the assistant field, strongly mediate the low entropy mode.

\begin{figure}[htbp]
\centering
\begin{minipage}[t]{0.48\textwidth}
\centering
\includegraphics[width=\textwidth]{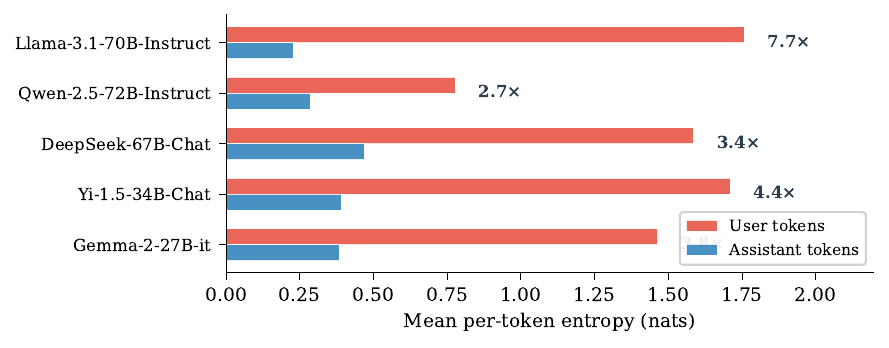}
\end{minipage}
\hfill
\begin{minipage}[t]{0.50\textwidth}
\centering
\includegraphics[width=\textwidth]{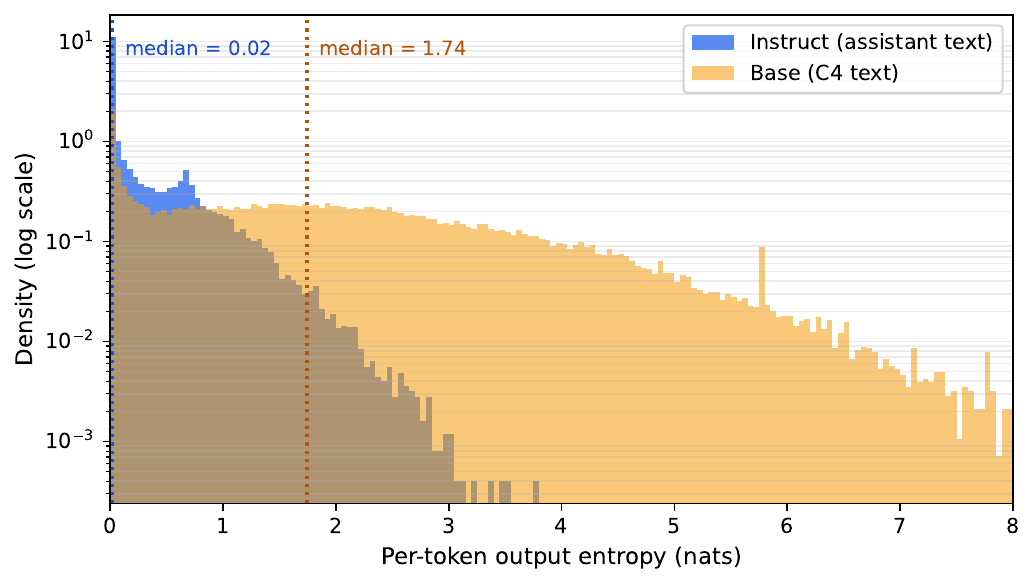}
\end{minipage}
\caption{\textbf{Post-trained models are low-entropy only in the Assistant role
and on their own text.} \textbf{Left:} Mean per-token output entropy by role,
across five instruct models each evaluated on its own multi-turn conversations.
Assistant-turn entropy is significantly lower than user-turn entropy.
\textbf{Right:} Full per-token entropy distributions for Llama-3.1-70B-Instruct
on its own chat outputs (blue, median 0.02 nats) versus on C4 web text with no
chat template (orange, median 1.74 nats). The two distributions are mostly non-overlapping.}

\label{fig:entropy_combined}
\end{figure}

To measure this, we simulated multi-turn conversations (Qwen-2.5-1.5B-Instruct playing the user, the target model playing the assistant, seeded from 30 open-ended topics; see Appendix~\ref{sec:prompts_appendix}), grouped tokens by speaker (user vs.\ assistant) and measured per-token output entropy $\Ent = -\sum_v p_v \log p_v$ (in nats). Figure~\ref{fig:entropy_combined} (left) reports this across five instruct-tuned models.

In Figure~\ref{fig:entropy_combined} (right) we also compared the full distribution of entropies between on-policy chat text and off-policy C4 data (English-language web text \citet{raffel2020c4}). On-policy, the median is near zero and nearly all tokens fall below 1 nat; off-policy on C4, the distribution is broader and with a much higher median; the two distributions are mostly non-overlapping.

It is clear that output entropy is lower during on-policy Assistant generations than user turns or pretraining-style data. This observation raises several questions. Is this effect due to the context containing on-policy generations, or due primarily to the model writing outputs for the default Assistant persona? More generally, can output entropy be used as a sign of implicit on-policy recognition by the model? If so, how does the model differentiate on-policy text from memorized data, which also yields low output entropy? The rest of this section explores these questions.

\subsection{The effects of on-policy vs. off-policy contexts on output entropy}
\label{sec:source_comparison}

\begin{figure}[htbp]
\centering
\includegraphics[width=0.85\textwidth]{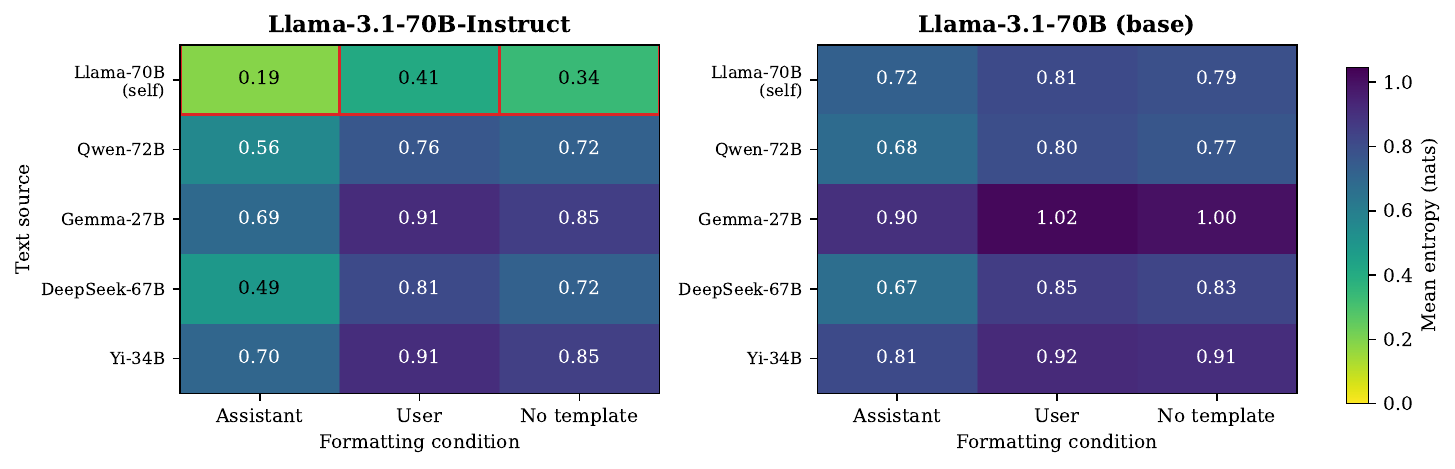}
\caption{\textbf{Self-generated text and Assistant formatting independently
lower output entropy; the base model shows neither effect.} Mean per-token
entropy (nats) when Llama-3.1-70B reads responses to the same 20 prompts from
itself and four other instruct models (rows), under three formatting conditions
(columns). \textbf{Left:} Instruct evaluator. The self row (boxed) is well
below all others in every column, with the largest gap in the Assistant
condition. \textbf{Right:} Base evaluator on the same texts. The self row is no
longer distinctive and formatting has only a modest effect.}
\label{fig:source_comparison}
\end{figure}

We evaluated Llama-3.1-70B-Instruct on responses generated by itself and four other frontier instruct models (Qwen-2.5-72B-Instruct, Gemma-2-27B-it, DeepSeek-67B-Chat, Yi-1.5-34B-Chat), all answering the same 20 prompts (Appendix~\ref{sec:prompts_appendix}), in three formatting conditions: as part of the Assistant turn, as part of the user turn, and using no chat template (Figure~\ref{fig:source_comparison}).

The instruct-tuned model produces substantially lower-entropy outputs when exposed to its own generations than to off-policy generations. The user/Assistant formatting amplifies this: Assistant turns decrease entropy even further. These results indicate that both on-policy text and being on the Assistant turn independently contribute to low-entropy outputs. The base model, by contrast, shows neither gap: it processes self-generated text at nearly identical entropy to text generated by other models, and shows only a modest dependence on formatting.

\begin{figure}[htbp]
\centering
\includegraphics[width=0.6\textwidth]{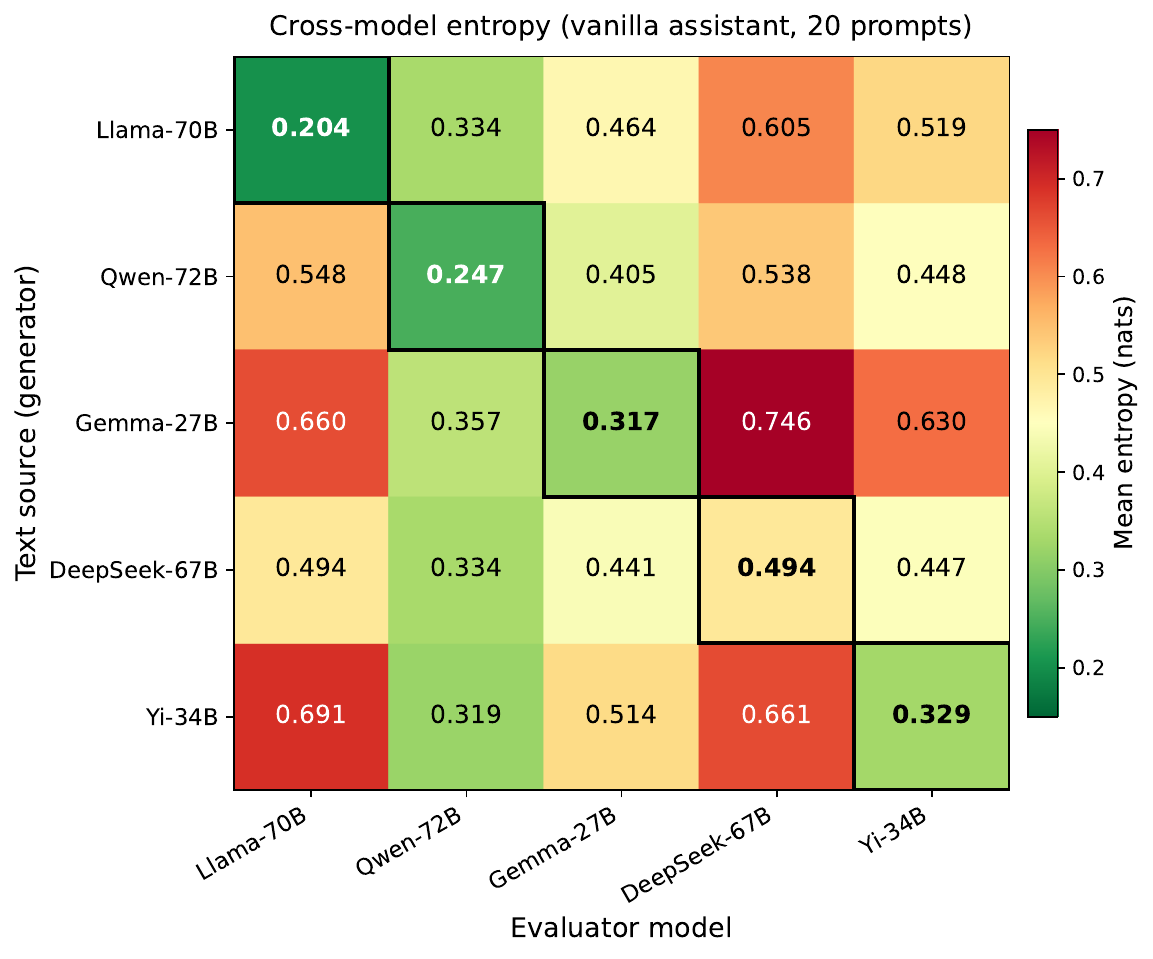}
\caption{\textbf{Every model has lower entropy while processing its own text than while processing any other
model's.} Mean per-token entropy (nats) for five instruct models in the
Assistant format, each evaluated on responses to the same 20 prompts generated
by every model in the suite. Rows: generator; columns: evaluator. The diagonal (evaluator = generator) is the column minimum in every column.}
\label{fig:cross_model}
\end{figure}

This experiment could in principle be confounded if Llama-70B's outputs are systematically different from other models' outputs in a way that admits lower-entropy predictions. To address this, we repeated the analysis reciprocally across all pairs of five different models. The effect is bidirectional: each model reads its own generations at lower entropy than any other model's (i.e. the diagonal is the column minimum in every column in Figure~\ref{fig:cross_model}). We take this as evidence that entropy reduction is driven, at least in part, by inputs generated from the model's own policy; one implication of this is that models can, at least implicitly, recognize their own generations, consistent with self-preference bias documented in LLM-as-judge evaluations~\citep{panickssery2024llm, wataoka2024selfpreference, ackerman2024inspection}.

\subsection{Model size and training stage}
\label{sec:size_and_stage}

Having established this self-generation recognition effect, we asked how it depends on model size and on different post-training algorithms.

\begin{figure}[htbp]
\centering
\includegraphics[width=\textwidth]{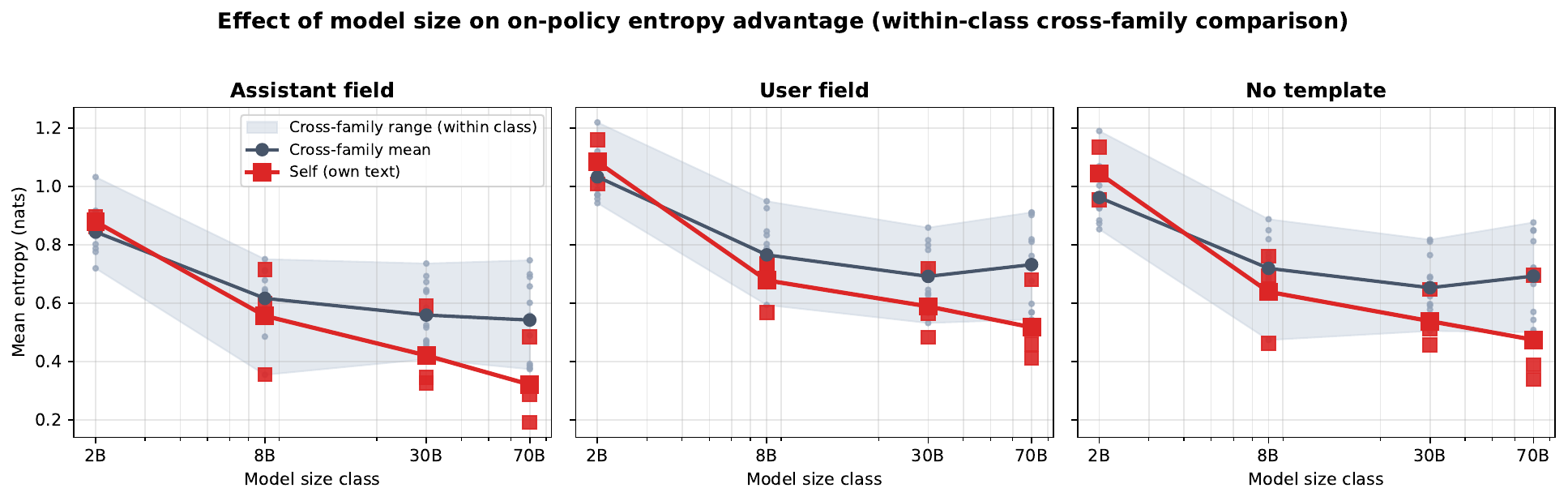}
\caption{\textbf{Self-recognition grows monotonically with model size.} Self
entropy (red) and cross-family entropy (grey band: range across the
Figure~\ref{fig:cross_model} suite; dark grey: mean) for instruct
models at each of four size classes ($\sim$2B, 8B, 30B, 70B). Panels: the three
formatting conditions. The self-advantage---the gap between red and grey---is
near zero at 2B and reaches 0.1--0.4 nats at 70B.}
\label{fig:size_effect}
\end{figure}

\paragraph{Size.} We ran the same cross-family comparison as above against the Figure~\ref{fig:cross_model} suite for two or three instruct models at each of four size classes: $\sim$2B (Gemma-2-2B-it, Qwen-2.5-1.5B-Instruct), $\sim$8B (Llama-3.1-8B-Instruct, Qwen-2.5-7B-Instruct, OLMo-3-7B-Instruct), $\sim$30B (Gemma-2-27B-it, Yi-1.5-34B-Chat, Qwen-2.5-32B-Instruct), and $\sim$70B (Llama-3.1-70B-Instruct, Qwen-2.5-72B-Instruct, DeepSeek-67B-Chat). The on-policy entropy reduction effect grows monotonically with size (Figure~\ref{fig:size_effect}): it is essentially absent at 2B (where self entropy is roughly equal to, or slightly higher than, entropy on the cross-family suite), and reaches 0.1--0.4~nats at 70B, which shows the largest effect.

\begin{figure}[htbp]
\centering
\includegraphics[width=\textwidth]{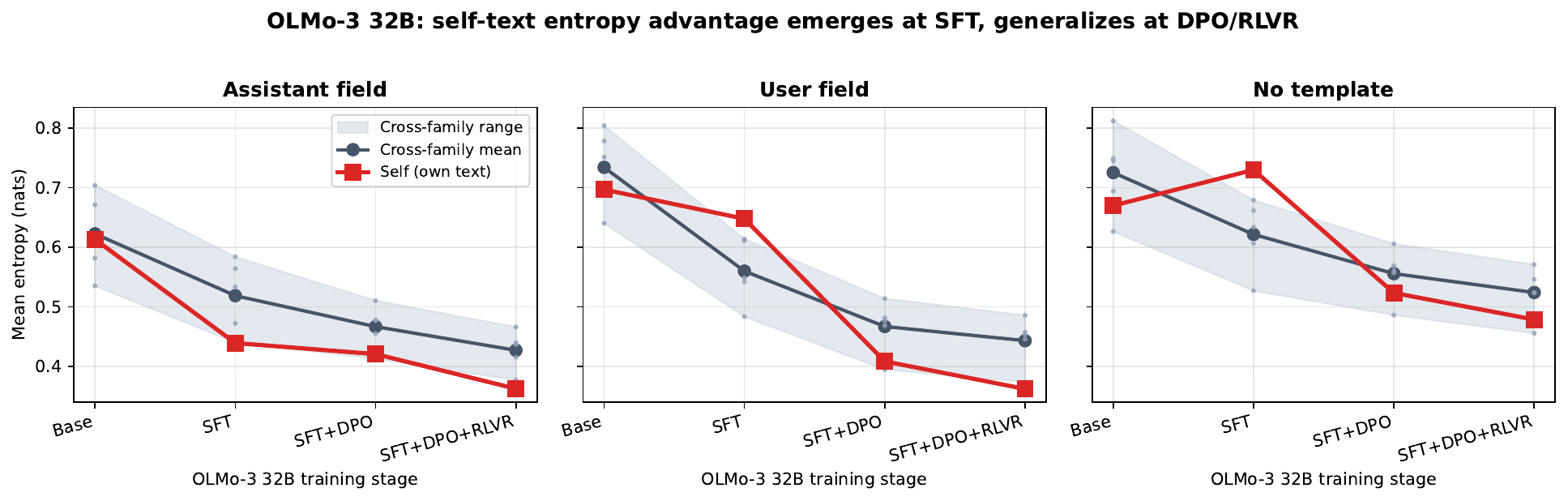}
\caption{\textbf{Effects of post-training stages on self-recognition} Self entropy (red) and cross-family entropy (grey band:
range; dark grey: mean) for OLMo-3-32B at four post-training checkpoints (Base
$\to$ +SFT $\to$ +DPO $\to$ +RLVR), evaluated on its own generations versus the
Figure~\ref{fig:cross_model} suite of other models' generations. Panels: the three formatting conditions.
The base model shows no self/other gap. SFT pulls same-model entropy below other-model entropy only in the Assistant condition
(and \emph{above} cross elsewhere). DPO brings  same-model entropy below other-model entropy in all three conditions;
RLVR modestly widens the gap.}
\label{fig:training_stages}
\end{figure}

\paragraph{Training stage.} OLMo-3 (32B) provides checkpoints at different stages of post-training: the same base model trained through SFT (supervised finetuning), then DPO (direct policy optimization), then RLVR (reinforcement learning with verifiable rewards). We measured each checkpoint's entropy on its own generations (self) versus on the Figure~\ref{fig:cross_model} suite (cross-family). The self-recognition effect is essentially absent in the base model. It appears at the SFT stage, but \emph{only in the Assistant field}: in the user and no-template conditions, the SFT checkpoint reads its own text at \emph{higher} entropy than the cross-family suite. DPO generalizes the recognition to those two conditions, pulling self-entropy below the cross-family mean everywhere. RLVR then further widens the gap in all three conditions, most strongly outside the assistant field (Figure~\ref{fig:training_stages}). SFT thus installs a role-gated form of on-policy recognition (the assistant role marker triggers it), DPO detaches the recognition from the role marker, and RLVR sharpens the overall effect.

The interpretation of these results is nuanced. The success of SFT and DPO  (two off-policy training methods) at engendering the self-recognition effect suggests that, perhaps surprisingly, on-policy training  is \emph{not} required to enable on-policy recognition. This suggests that the more relevant features of post-training may be the processes by which the data are selected (e.g. based on outcome or human feedback), and/or the specialization of training only on Assistant turns. For instance, consistent training on  Assistant outputs could increase the model's confidence on Assistant turns, which could lead to a cascading entropy reduction effect when combined with the base-model mechanism that ties low-surprise inputs to low-entropy outputs (Section~\ref{sec:autoregressive}).

\subsection{Effects of persona on self-recognition}
\label{sec:persona}

A natural question is whether the entropy decrease on self-written text is tied to the Assistant character's outputs, or if it applies to on-policy self-generated text presented outside the Assistant turn. Two experiments shed light on this question.

\begin{figure}[htbp]
\centering
\includegraphics[width=0.92\textwidth]{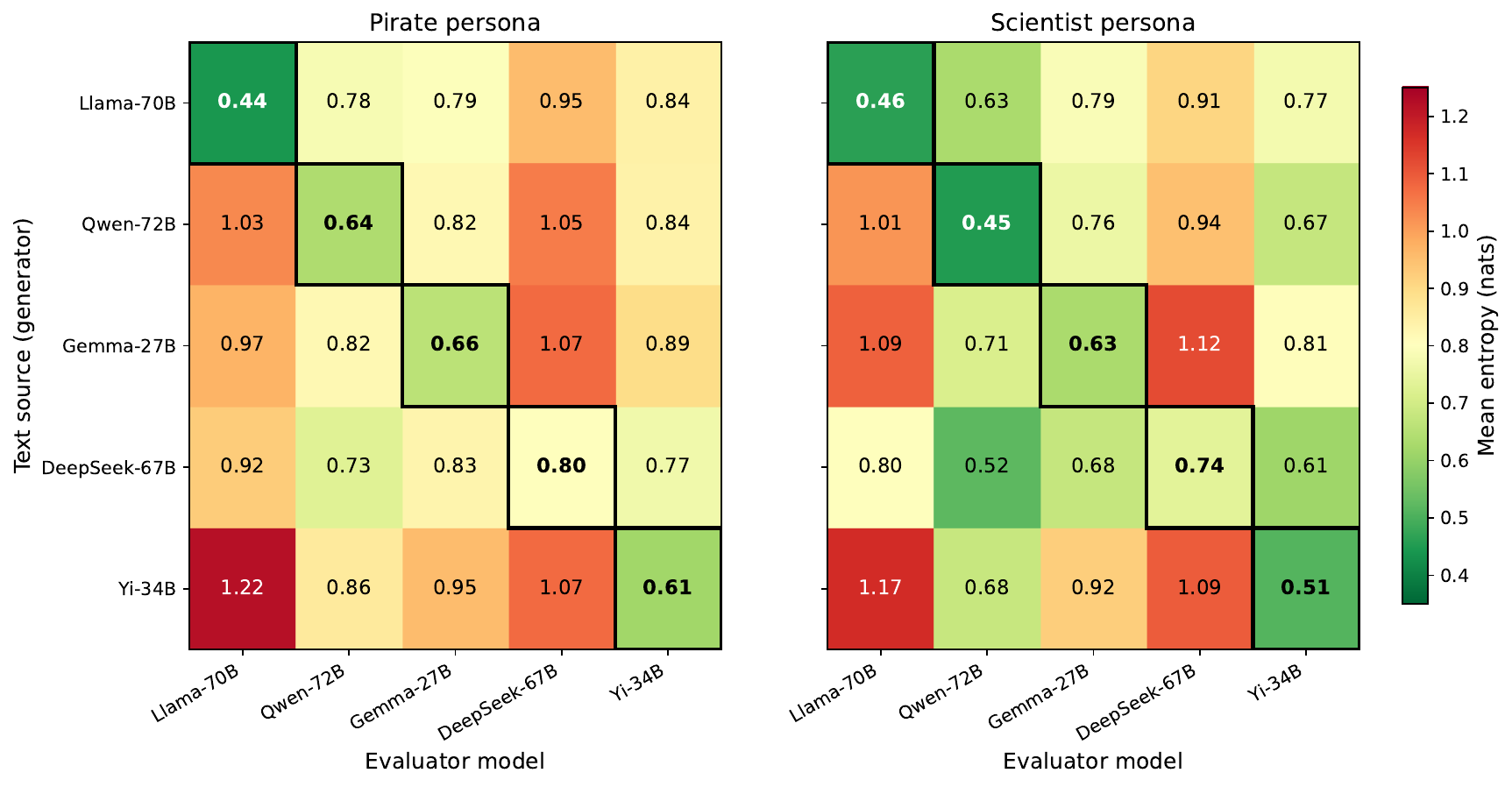}\\[0.6em]
\includegraphics[width=0.88\textwidth]{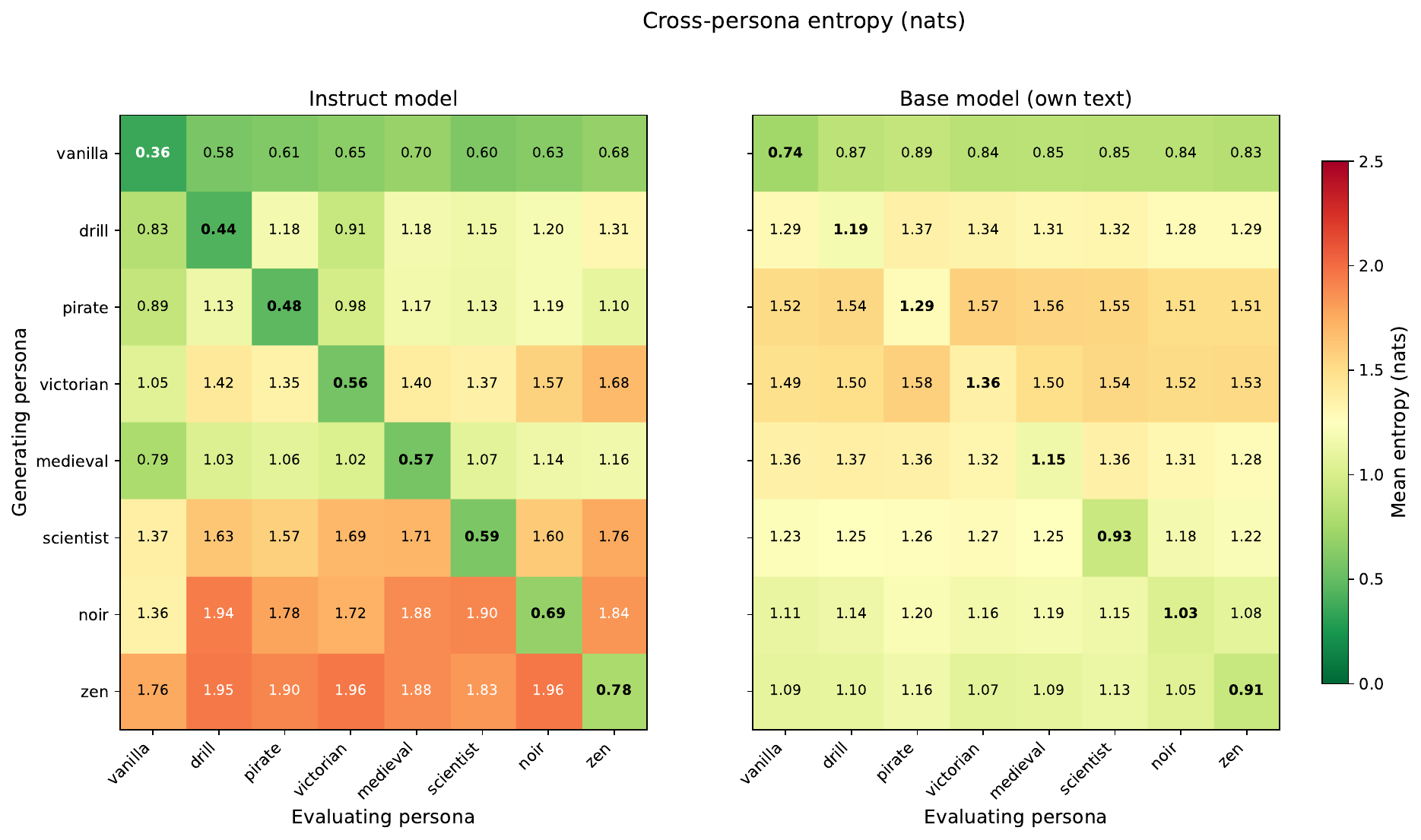}
\caption{\textbf{Effects of system-prompted persona on self-recognition.}
\textbf{Top:} The cross-model experiment of Figure~\ref{fig:cross_model}
repeated with all five models system-prompted as ``pirate'' (left) or
``scientist'' (right). The diagonal remains the column minimum in both cases,
indicating that same-model generations yield lower output entropy than
other-model generations regardless of which persona is used. \textbf{Bottom:} A
single model (Llama-3.1-70B-Instruct) generating under one of eight personas
(rows) and evaluated under another (columns). Entropy is lowest along the
matched-persona diagonal, and lowest of all when both generator and evaluator
use the default Assistant persona. The base model (right) shows a much
weaker effect.}
\label{fig:persona}
\end{figure}

First, we repeated the cross-model comparison of Figure~\ref{fig:cross_model} after system-prompting every model to adopt the same non-default persona, ``pirate'' or ``scientist,'' fixed across generator and evaluator. As above, self-generated text produces lower entropy outputs than text generated by other models in the same system-prompted persona, regardless of what persona that is (Figure~\ref{fig:persona}, top).

Second, we varied the system-prompted persona \emph{within a single model} (Llama-3.1-70B-Instruct), generating text under one persona and evaluating it under another. The lowest entropy occurs when generating and evaluating persona match, and an even lower entropy when both are the default Assistant (Figure~\ref{fig:persona}, bottom). Moreover, we find that this effect is post-training specific.

Taken together, text generated by a model in its default Assistant persona produces the lowest output entropy when input into that same model. Deviating from the Assistant persona in both the generating and evaluating models imposes a small entropy increase, using mismatched personas for the two models imposes a larger one, and using a different model altogether imposes an increase of variable size depending on the models in question (but consistently positive). These results suggest an account where both \emph{likelihood} (according to the model) and \emph{familiarity} (amount of weight during training) of the context are both contributors to low-entropy output distributions.

\subsection{The model represents its own entropy internally}
\label{sec:representations_brief}

\begin{figure}[htbp]
\centering
\includegraphics[width=0.95\textwidth]{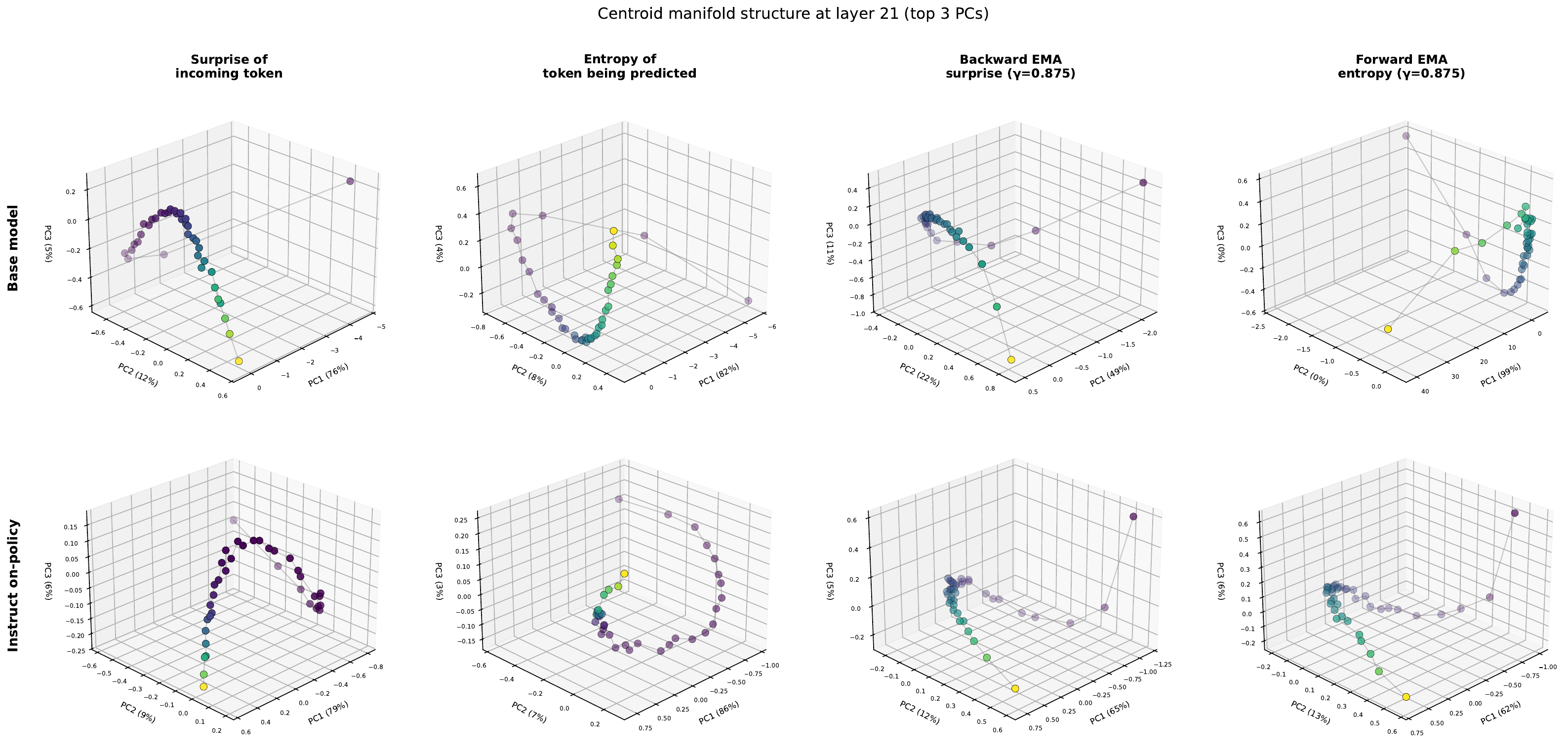}
\caption{\textbf{Internal representations of entropy and surprise.} Hidden
states at layer 21 are binned by feature value (color) and averaged within
each bin; the resulting centroids are projected onto their top three principal
components. Columns: four features (surprise of the incoming token, entropy of
the predicted token, and backward and forward EMA of entropy). Rows: base
model on web text (top) and instruct model on on-policy generations (bottom). In both cases
the centroids trace structured one-dimensional curves, but the base and on-policy
curves occupy nearly orthogonal subspaces
(Appendix~\ref{sec:orthogonal_appendix}).}
\label{fig:manifold_4feat}
\end{figure}

Entropy is an informative indicator of self-recognition, but as it is a property of the output distribution, this raises the question of whether the model maintains any \emph{internal} representations related to or causally upstream of its entropy. To answer this, we binned hidden states at layer 21 by four features---surprise of incoming token, entropy of the token being predicted, and backward and forward exponential moving average (EMA) of entropy---and averaged within each bin to produce a centroid for each bin. We repeated this for both the base model (on pretraining-like text) and the instruct-tuned model (on on-policy text). These centroids traced ordered one-dimensional curves in activation space (Figure~\ref{fig:manifold_4feat}).

We observed significant differences in the base model and on-policy instruct-tuned model representations of these quantities. Mean cosine similarity of centered centroids at matched nat values hovered near zero across all 80 layers. Linear centered kernel alignment (CKA) yielded modest values of 0.2--0.5, indicating that the manifold shapes are partially preserved but rotated into orthogonal directions (Appendix~\ref{sec:orthogonal_appendix}). This comparison varies several factors at once: it pits a base model
against an instruct model, web text against chat-formatted text, and
off-policy text against on-policy text. To isolate the last factor, we
ran a control with the model and chat formatting held fixed: we
computed centroids for the same instruct-tuned model reading
multi-turn conversations in the same format, varying only whether the
assistant turns were its own on-policy generations or generations
swapped in from a different model (Qwen-2.5-3B-Instruct). Even with
the model and formatting held constant, the centroid subspaces
remained nearly as orthogonal as in the original base-versus-on-policy
comparison. This suggests that the distinct representations are differentiated primarily by whether they apply to on- vs. off-policy text, regardless chat formatting. This phenomenon is investigated
further in Appendix~\ref{sec:orthogonal_appendix}.

However, while these representations are easily decoded, many of them are not causal. Notably, steering towards any centroid (relative to the mean) in the on-policy entropy manifold shifts output entropy by only marginal amounts across its full bin range (Appendix~\ref{sec:entropy_steering_appendix}). However, the internal surprise representation \emph{does} have a causal effect on output entropy, discussed in the next section.

\subsection{The effect of input surprise on output entropy}
\label{sec:autoregressive}

How do on-policy generations provided as input result in low-entropy outputs? One hypothesis is that the model computes the surprise of each input token relative to its prior predictions, and decreases output entropy in contexts with low surprise. To test this, we first analyzed the dynamics of output entropy over the course of extended on-policy generations; if entropy is reduced when low-surprise tokens are encountered, we should expect entropy to decrease over the course of a sample when the sampling is sufficiently deterministic (low temperature).

\begin{figure}[htbp]
\centering
\includegraphics[width=0.5\textwidth]{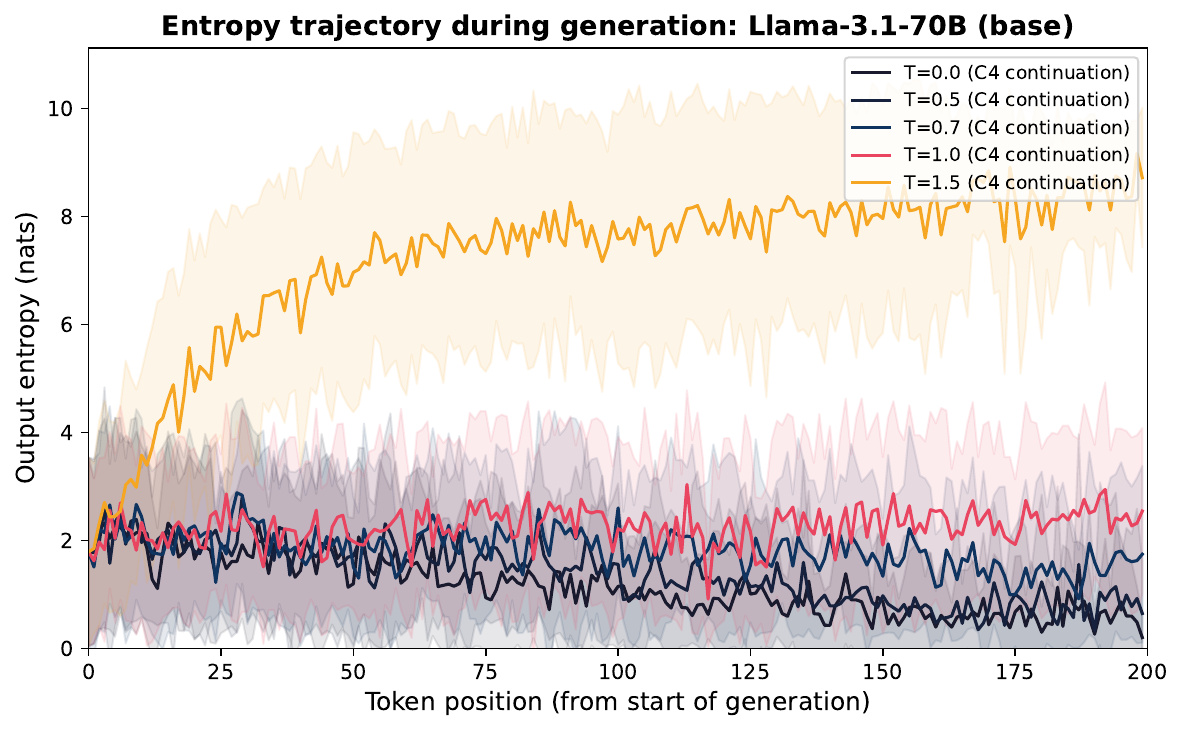}
\caption{\textbf{Output entropy decreases during low-temperature generation,
even in the base model.} Per-position output entropy during autoregressive
generation from Llama-3.1-70B base (20 web text continuations per temperature;
shaded region $\pm 1$ std). At $T{=}1$ entropy is approximately flat over the
sequence; at lower temperatures it decays as low-surprise tokens accumulate in
the context.}
\label{fig:entropy_traj_base}
\end{figure}

We found that this effect is indeed present, even in the base model. Its output entropy decreased over time when sampling with low temperatures, and increased when sampling with a high temperature (Figure~\ref{fig:entropy_traj_base}). We will see shortly that this mechanism is greatly amplified in the instruct-tuned model when on-policy.

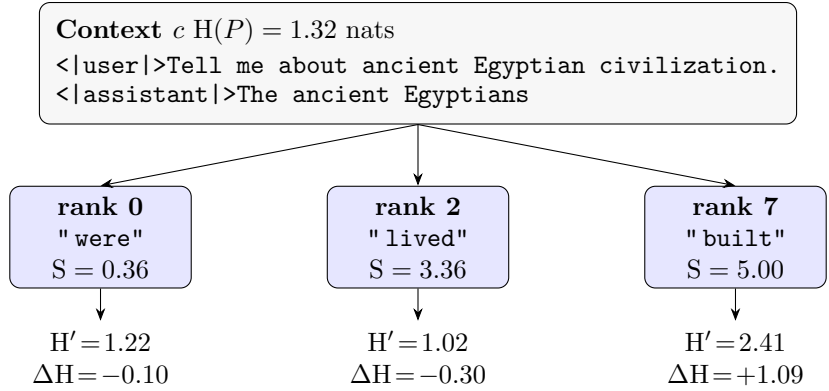
\begin{figure}[htbp]
\centering
\begin{tikzpicture}[font=\small]
  \node[draw, rounded corners, inner sep=6pt, fill=black!3, minimum width=8.0cm, align=left] (ctx) at (0, 0) {
    \textbf{Context $c$} \hfill $\Ent(P) = 1.32$ nats \\[1pt]
    \texttt{<|user|>Tell me about ancient Egyptian civilization.}\\[-1pt]
    \texttt{<|assistant|>The ancient Egyptians}
  };
  \node[draw, rounded corners, fill=blue!10, align=center, minimum width=2.4cm] (tk1) at (-4.2, -2.3) {\textbf{rank 0} \\ \texttt{``\,were''} \\ $\Surp = 0.36$};
  \node[draw, rounded corners, fill=blue!10, align=center, minimum width=2.4cm] (tk2) at (0,    -2.3) {\textbf{rank 2} \\ \texttt{``\,lived''} \\ $\Surp = 3.36$};
  \node[draw, rounded corners, fill=blue!10, align=center, minimum width=2.4cm] (tk3) at (4.2,  -2.3) {\textbf{rank 7} \\ \texttt{``\,built''} \\ $\Surp = 5.00$};
  \draw[-{Stealth}] (ctx.south) -- (tk1.north);
  \draw[-{Stealth}] (ctx.south) -- (tk2.north);
  \draw[-{Stealth}] (ctx.south) -- (tk3.north);
  \node[align=center] (o1) at (-4.2, -3.9) {$\Ent' \!=\! 1.22$ \\ $\Delta\Ent \!=\! -0.10$};
  \node[align=center] (o2) at (0,    -3.9) {$\Ent' \!=\! 1.02$ \\ $\Delta\Ent \!=\! -0.30$};
  \node[align=center] (o3) at (4.2,  -3.9) {$\Ent' \!=\! 2.41$ \\ $\Delta\Ent \!=\! +1.09$};
  \draw[-{Stealth}] (tk1.south) -- (o1.north);
  \draw[-{Stealth}] (tk2.south) -- (o2.north);
  \draw[-{Stealth}] (tk3.south) -- (o3.north);
\end{tikzpicture}
\caption{\textbf{Protocol for measuring the effect of input
surprise on output entropy} (illustrated with one example chat context,
Llama-3.1-70B-Instruct). Given the fixed context shown, the model produces an
output distribution $P$ with entropy $\Ent$. We append a single token $w$
drawn from a specified rank of $P$ and read off the entropy $\Ent'$ of the
resulting next-position distribution. Three of the twenty ranks swept are
shown: the argmax (\texttt{were}), another high-ranking alternative (\texttt{lived}), and a
higher-surprise but still plausible continuation (\texttt{built}). Sweeping all
possible ranks across many contexts produces the data fit in the subsequent
Figure~\ref{fig:single_step}.}
\label{fig:single_step_schematic}
\end{figure}

To further test the hypothesis that surprise contributes to output entropy, we conducted a more controlled experiment. Given a fixed context $c$ the model yields an output distribution $P$ with entropy $\Ent$; we intervened by appending one token $w$ drawn from $P$ and reading off the model's new output distribution $P'$ and its entropy $\Ent'$. Since the context is fixed, the change $\Ent' - \Ent$ depends only on the appended token $w$, and we can sweep the sampled $w$ across the quantiles of the $P$ distribution. Figure~\ref{fig:single_step_schematic} illustrates the protocol with an example.

We repeated this intervention across three experimental conditions: (i) 20 \emph{chat contexts}---an instruct model (Llama-3.1-70B) with its chat template applied to the twenty prompts of Appendix~\ref{sec:prompts_appendix}, measured at the final prompt position (on-policy entry into assistant generation); (ii) 50 \emph{web text contexts} on the instruct model---with \emph{no} chat template, reading passages of web text; and (iii) 50 \emph{web contexts}---the corresponding base model reading the same web text.

In on-policy generations, the expected surprise of a given input token is equal to the entropy of the prior output distribution. Thus, it is helpful to quantify the degree of surprise of a token relative to expectations, by defining the \emph{relative excess surprise} $(\Surp - \Ent)/\Ent$, where $\Surp$ and $\Ent$ denote surprise and entropy, respectively. We also define the \emph{relative entropy change} from one token position's output distribution to the next, $\Delta \Ent / \Ent = (\Ent' - \Ent)/\Ent$.

We found that the relative entropy change had an approximately linear relationship to the relative excess surprise. We fit a linear relationship as:
\begin{equation}
\frac{\Delta\Ent}{\Ent} \;\approx\; a \cdot \frac{\Surp - \Ent}{\Ent} \;+\; \beta
\label{eq:feedback}
\end{equation}
The sensitivity $a$ to relative excess surprise was relatively stable across all three conditions. However, the intercept $\beta$ differed sharply: strongly negative with chat formatting, and near zero with both the base model and the instruct model outside of chat formatting. In other words, under chat formatting output entropy drops even when the token has surprise in line with expectations, whereas outside it, entropy drifts only in response to excessively surprising or unsurprising tokens.

\begin{figure}[htbp]
\centering
\includegraphics[width=0.7\textwidth]{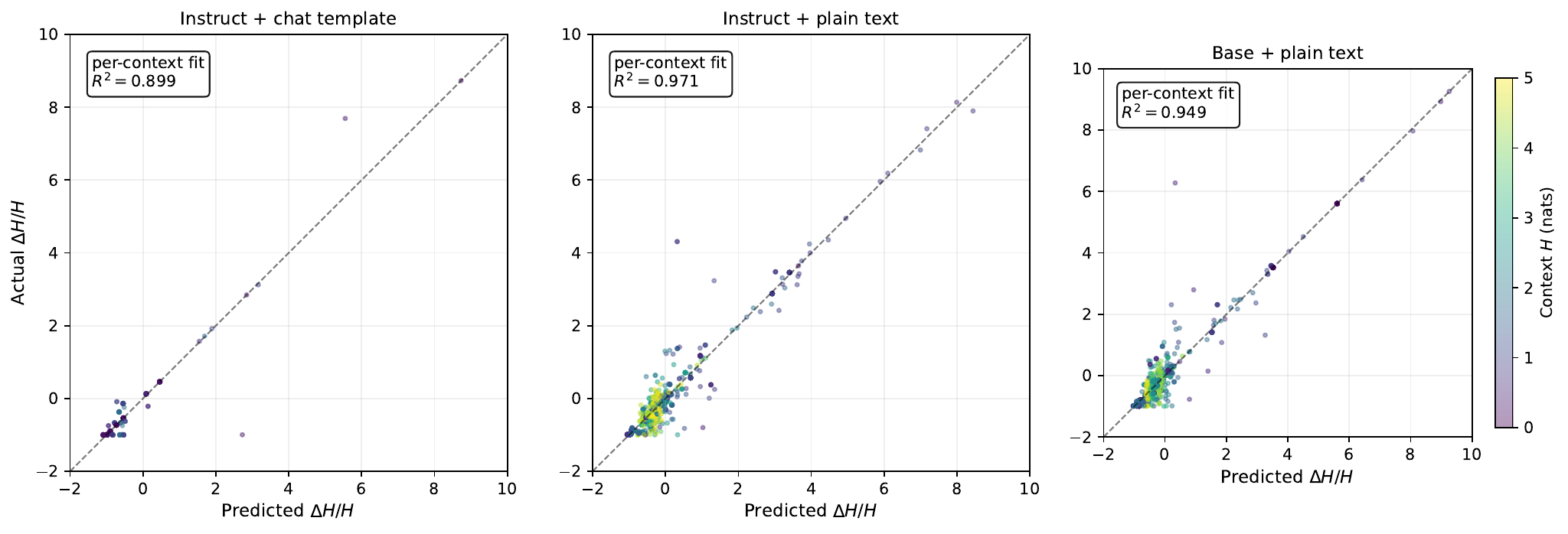}
\caption{\textbf{Relationship between input surprise and output entropy across
conditions.} Predicted versus actual relative entropy change $\Delta\Ent/\Ent$
under the linear fit of Equation~\ref{eq:feedback}; each point is one
(context, appended token) pair. The fitted sensitivity $a$ is similar across
all three conditions, while the intercept $\beta$ is strongly negative only in
the chat condition---meaning that under chat formatting, output entropy
decreases even when the appended token is exactly as surprising as expected.}
\label{fig:single_step}
\end{figure}

\begin{figure}[htbp]
\centering
\includegraphics[width=0.8\textwidth]{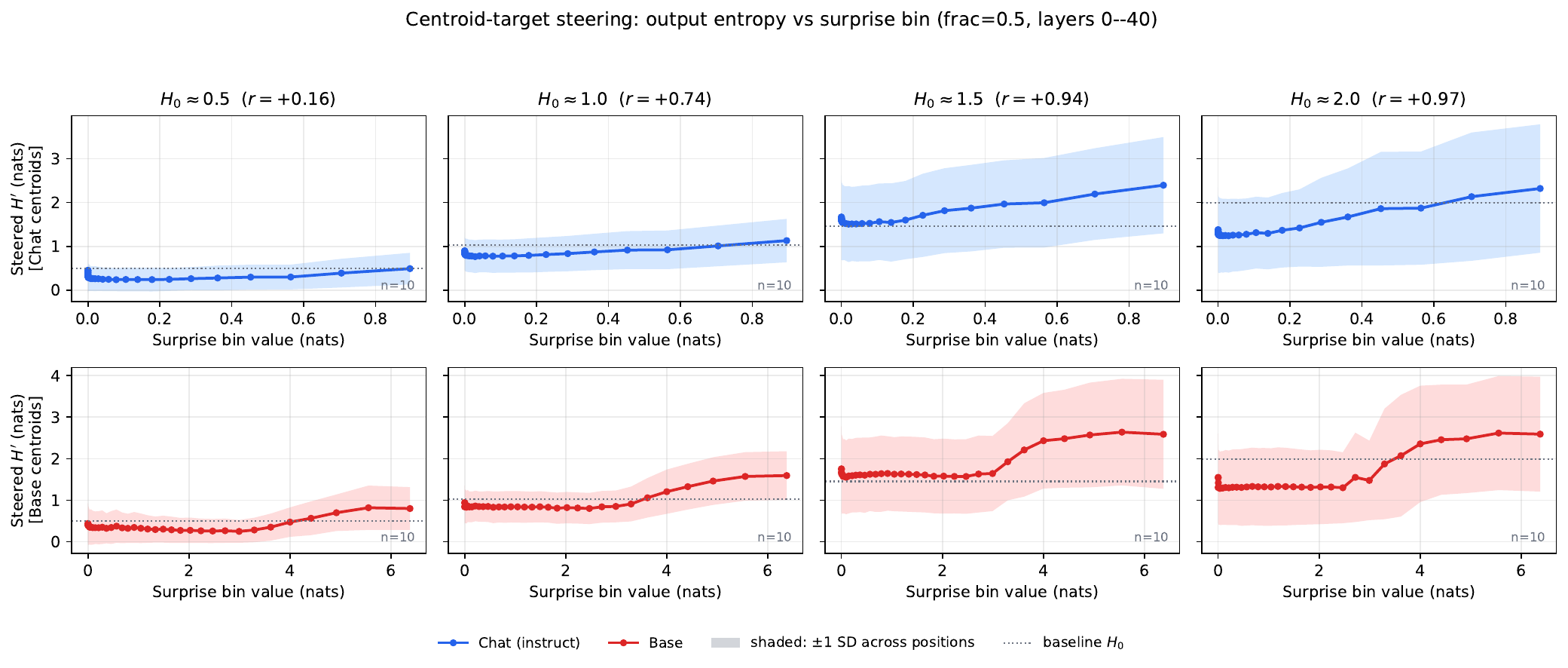}
\caption{\textbf{Steering along the surprise representation modulates output
entropy.} Output entropy after steering layer 0--39 activations toward each
surprise-centroid bin (at half the bin's displacement from the mean; ten
contexts per panel). Panels correspond to different levels of baseline entropy
$\Ent_0$. Top row: on-policy centroids; bottom row: base-model centroids.
Solid lines show the per-bin mean across the ten positions; shaded bands show
$\pm 1$ standard deviation across positions. Steering toward low-surprise bins
lowers output entropy and steering toward high-surprise bins raises it.}
\label{fig:steer_by_entropy}
\end{figure}

What is the mechanism by which excess surprise drives these changes in output entropy? In Section~\ref{sec:representations_brief} we identified activation-space representations that track the surprise of the incoming token (Figure~\ref{fig:manifold_4feat}). A natural hypothesis is that those representations are the causal substrate of Equation~\ref{eq:feedback}. If so, intervening on the activations along these directions should reproduce the entropy response we saw from natural token substitutions. For a given context with baseline entropy $\Ent_0$, we steered toward each centroid bin at half its displacement from the mean, across layers 0--39 (Figure~\ref{fig:steer_by_entropy}). We found that steering in the vector directions associated with higher surprise values resulted in higher output entropy.

\section{Pre-response Planning and Intent Continuity}
\label{sec:preplanning}

Our results thus far suggest that post-trained models implicitly recognize their own text, that this recognition is reflected in the entropy of their output distribution, and that it is based (at least in part) on an estimate of the surprise of inputs in the context. Section~\ref{sec:entropy} focused on entropy at the token level; in this section we examine a complementary form of uncertainty that we call \emph{semantic entropy}: the spread of the distribution over which \emph{topic} the model is about to generate a response on, rather than which next \emph{token}. We show that instruct models collapse this semantic uncertainty at response time (committing to a single topic before their first output token), and that disrupting that commitment raises token-level output entropy.

We test the model's semantic entropy using eight pairs of prompts (Table~\ref{tab:kv_domains}). Each pair consists of an underspecified prompt (``Think of a food and explain why you find it interesting'') and a specific prompt (``Describe haggis and explain why you find it interesting'') with matched token counts and compatible topics. We also obtained prefills using the model's own temperature zero output from the specific prompt, truncated to a few tokens.

\begin{table}[H]
\centering
\caption{Domain-matched prompt pairs used throughout this section.}
\label{tab:kv_domains}
\footnotesize
\setlength{\tabcolsep}{4pt}
\begin{tabular}{llll}
\toprule
\textbf{Domain} & \textbf{Underspecified} & \textbf{Specific} & \textbf{Prefill} \\
\midrule
food & Think of a food\ldots & Describe haggis\ldots & ``Haggis is'' \\
sport & Think of a sport\ldots & Tell me what makes hockey\ldots & ``Hockey is an'' \\
element & Think of a chemical element\ldots & Tell me about mercury\ldots & ``Mercury is a'' \\
art\_form & Think of an art form\ldots & I want to know about sculpture\ldots & ``Sculpture is a\ldots'' \\
technology & Think of a technology\ldots & Tell me about AI\ldots & ``Artificial intelligence\ldots'' \\
historical\_fig. & Think of a historical figure\ldots & Describe Genghis Khan\ldots & ``Genghis Khan'' \\
philosopher & Think of a philosopher\ldots & Explain Socrates\ldots & ``Socrates was a'' \\
invention & Think of an invention\ldots & Tell me about the telephone\ldots & ``The telephone is a\ldots'' \\
\bottomrule
\end{tabular}
\end{table}

\subsection{Implicit commitment reduces semantic entropy}
\label{sec:commitment}

Given an underspecified prompt (``Think of a food...''), a model with high semantic entropy spreads its output probability mass over many possible food choices, while a model with low semantic entropy commits to one (haggis, apples, pizza) before producing its first token. The base model is trained to model an unknown distribution in which many choices are plausible, and thus we would expect to exhibit high semantic entropy on such a prompt. An instruct model, on the other hand, may be more opinionated in its topic selection.

\begin{figure}[htbp]
\centering
\includegraphics[width=0.7\textwidth]{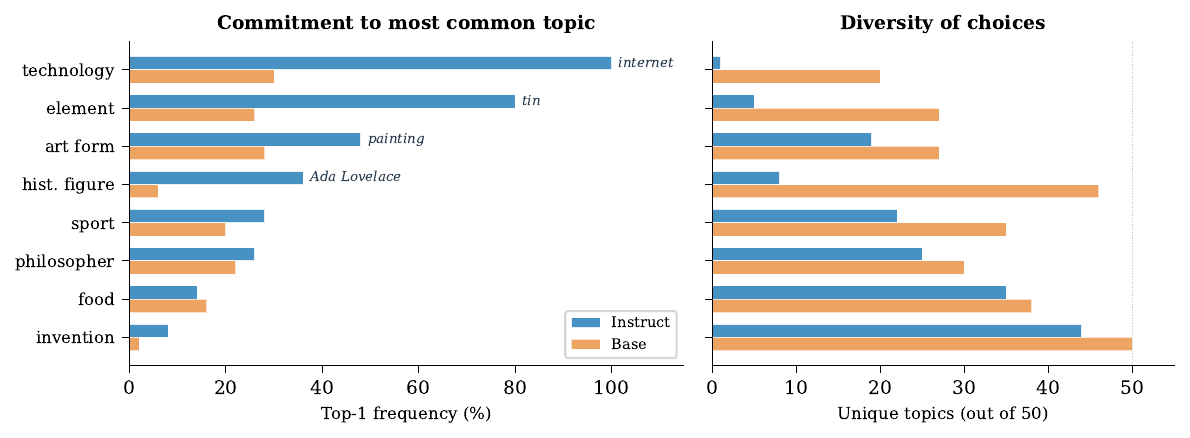}
\caption{\textbf{Topic commitment on underspecified prompts.} Fifty
completions sampled at $T{=}1$ from each of the eight underspecified prompts in
Table~\ref{tab:kv_domains}, for both the instruct and base models.
\textbf{Left:} Fraction of completions choosing the single most common topic.
\textbf{Right:} Number of distinct topics appearing across the fifty samples.
The instruct model's distribution over topics is substantially more concentrated than the base model on
both measures, in every domain.}
\label{fig:commitment}
\end{figure}

In Figure~\ref{fig:commitment}, we generated 50 completions at temperature 1 from each of the eight underspecified prompts, for both the instruct and base models, and recorded which specific topic appeared.

\subsection{Prefills inconsistent with intended generations result in increased output entropy}
\label{sec:crossover}

Here we show that subverting the model's implicit topic commitment by providing a prefill from a different topic resulted in an increase in output entropy, even when the prefill was the model's own generation.

We compared the results of using these prefills with the specific prompt (in which case they are on-policy) and with the general prompt (off-policy). We sampled ten generations per condition and measured entropy over tokens 6--300. The instruct model had higher entropy when the prefill was appended to the specific prompt (where it is on-policy) than when appended to the general prompt (where it is off-policy). Interestingly, we observed the \emph{reverse} effect in the base model, presumably because the specific-topic prefill narrows the otherwise unconstrained distribution of responses to the underspecified question.

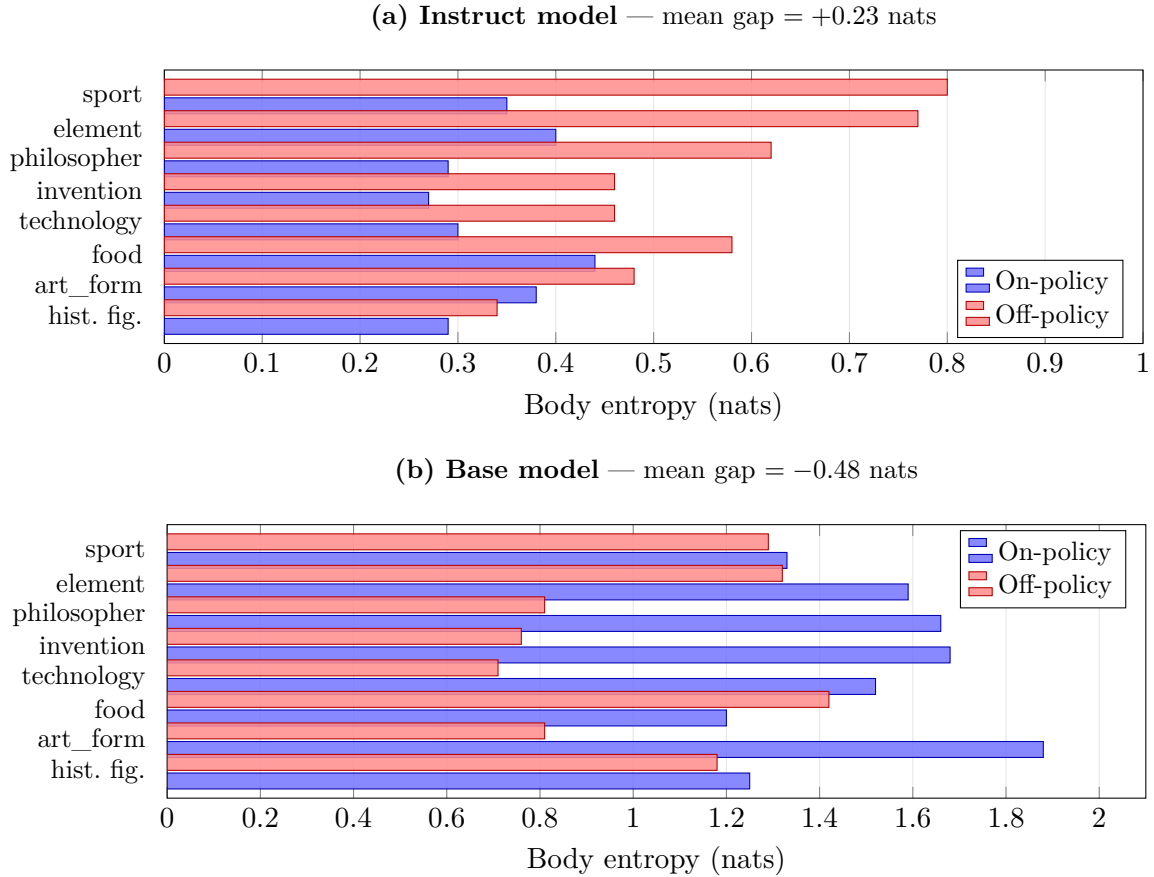
\begin{figure}[H]
\centering

\begin{tikzpicture}
\begin{axis}[
    width=0.88\textwidth,
    height=5.2cm,
    xbar=0pt,
    bar width=6pt,
    y dir=reverse,
    ytick={1,2,3,4,5,6,7,8},
    yticklabels={sport, element, philosopher, invention, technology, food, art\_form, hist.\ fig.},
    yticklabel style={font=\small},
    xmin=0, xmax=1.0,
    xlabel={Body entropy (nats)},
    legend style={at={(0.98,0.02)}, anchor=south east, font=\small},
    legend cell align=left,
    every axis plot/.append style={fill opacity=0.85},
    ytick style={draw=none},
    xmajorgrids=true,
    grid style={gray!20},
    enlarge y limits=0.12,
    title={\textbf{(a) Instruct model} --- mean gap $= +0.23$ nats},
    title style={at={(0.5,1.05)}, font=\small},
]

\addplot[fill=blue!55, draw=blue!70!black, bar shift=-3.5pt] coordinates {
    (0.35, 1) (0.40, 2) (0.29, 3) (0.27, 4) (0.30, 5) (0.44, 6) (0.38, 7) (0.29, 8)
};
\addplot[fill=red!45, draw=red!70!black, bar shift=+3.5pt] coordinates {
    (0.80, 1) (0.77, 2) (0.62, 3) (0.46, 4) (0.46, 5) (0.58, 6) (0.48, 7) (0.34, 8)
};

\legend{On-policy, Off-policy}
\end{axis}
\end{tikzpicture}

\vspace{0.2cm}

\begin{tikzpicture}
\begin{axis}[
    width=0.88\textwidth,
    height=5.2cm,
    xbar=0pt,
    bar width=6pt,
    y dir=reverse,
    ytick={1,2,3,4,5,6,7,8},
    yticklabels={sport, element, philosopher, invention, technology, food, art\_form, hist.\ fig.},
    yticklabel style={font=\small},
    xmin=0, xmax=2.1,
    xlabel={Body entropy (nats)},
    legend style={at={(0.98,0.98)}, anchor=north east, font=\small},
    legend cell align=left,
    every axis plot/.append style={fill opacity=0.85},
    ytick style={draw=none},
    xmajorgrids=true,
    grid style={gray!20},
    enlarge y limits=0.12,
    title={\textbf{(b) Base model} --- mean gap $= -0.48$ nats},
    title style={at={(0.5,1.05)}, font=\small},
]

\addplot[fill=blue!55, draw=blue!70!black, bar shift=-3.5pt] coordinates {
    (1.33, 1) (1.59, 2) (1.66, 3) (1.68, 4) (1.52, 5) (1.20, 6) (1.88, 7) (1.25, 8)
};
\addplot[fill=red!45, draw=red!70!black, bar shift=+3.5pt] coordinates {
    (1.29, 1) (1.32, 2) (0.81, 3) (0.76, 4) (0.71, 5) (1.42, 6) (0.81, 7) (1.18, 8)
};

\legend{On-policy, Off-policy}
\end{axis}
\end{tikzpicture}

\caption{\textbf{The effect of off-plan prefills on output entropy.} Mean body
entropy (tokens 6--300, ten generations per condition) when a specific-topic
prefill from Table~\ref{tab:kv_domains} follows the matching specific prompt
(on-policy, blue) versus the underspecified prompt (off-policy, red).
\textbf{(a)~Instruct model:} off-policy entropy exceeds on-policy entropy in
every domain, consistent with the prefill conflicting with the model's cached
intent. \textbf{(b)~Base model:} the pattern reverses, perhaps due to the prefill narrowing an otherwise
unconstrained continuation.}
\label{fig:entropy_cost}
\end{figure}

\subsection{Explicit prefill detection}
\label{sec:kv_patching}

The previous experiments showed that the instruct model stores topic intent, and that deviating from the planned topic increases its output entropy. Here, we show that Llama-3.1-70B also has a more \emph{explicit} self-recognition ability that enables it to report on deviations between artificially prefilled responses and its cached intent. We find that this capability is sensitive to the alignment between hidden activations representing intent at the user-token positions, and the content of the response, and that this comparison is computed immediately before the prefill detection verdict.

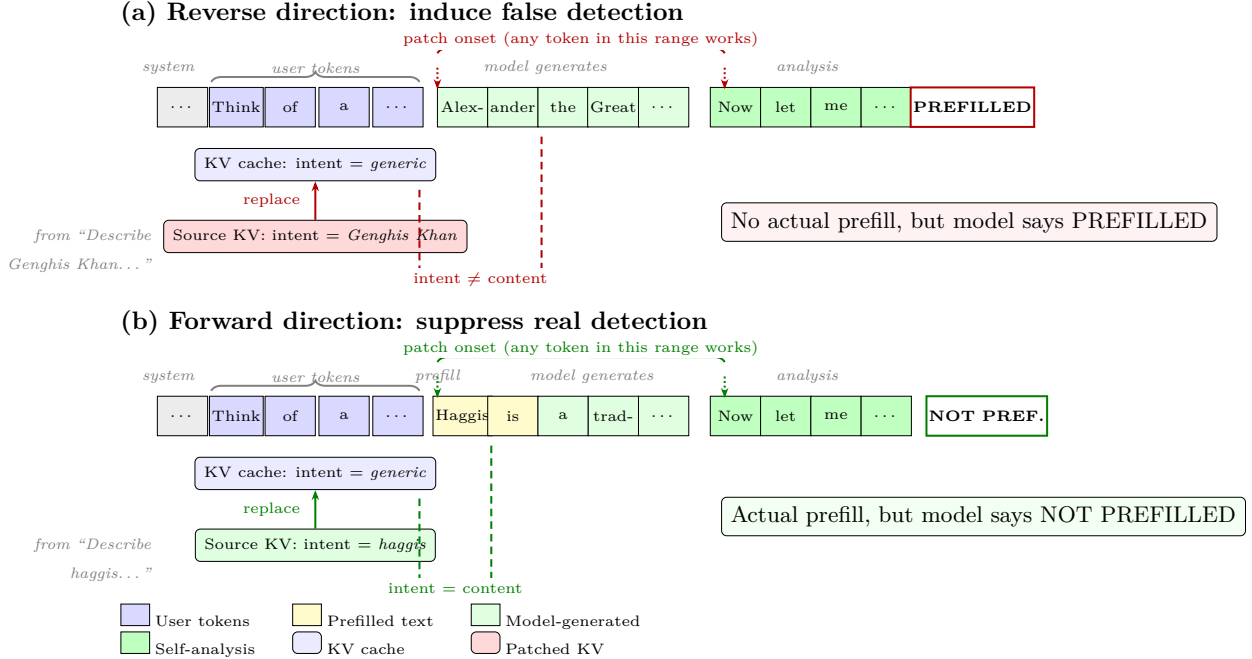
\begin{figure}[H]
\centering
\resizebox{\textwidth}{!}{%
\begin{tikzpicture}[
    x=1cm, y=1cm,
    tok/.style={draw, minimum height=0.55cm, minimum width=0.7cm, font=\tiny, inner sep=1pt},
    sys/.style={tok, fill=gray!15},
    usr/.style={tok, fill=blue!15},
    pre/.style={tok, fill=yellow!25},
    gen/.style={tok, fill=green!12},
    ana/.style={tok, fill=green!25},
    verdict/.style={tok, fill=white, thick},
    kvbox/.style={draw, fill=blue!8, rounded corners=2pt, minimum height=0.45cm, font=\tiny},
    kvpatch/.style={draw, fill=red!15, rounded corners=2pt, minimum height=0.45cm, font=\tiny},
    arr/.style={-{Stealth[length=4pt]}, thick},
    brace/.style={decorate, decoration={brace, amplitude=4pt, mirror}},
    bracetop/.style={decorate, decoration={brace, amplitude=4pt}},
]

\node[font=\small\bfseries, anchor=west] at (-1, 5.6) {\textbf{(a)} Reverse direction: induce false detection};

\node[font=\tiny\itshape, text=gray] at (-0.2, 4.85) {system};
\node[font=\tiny\itshape, text=gray] at (1.85, 4.85) {user tokens};
\node[font=\tiny\itshape, text=gray] at (5.05, 4.85) {model generates};
\node[font=\tiny\itshape, text=gray] at (8.7, 4.85) {analysis};

\node[sys] (s1) at (0, 4.3) {\ldots};
\node[usr] (u1) at (0.75, 4.3) {Think};
\node[usr] (u2) at (1.5, 4.3) {of};
\node[usr] (u3) at (2.25, 4.3) {a};
\node[usr] (u4) at (3.0, 4.3) {\ldots};
\node[gen] (g1) at (3.9, 4.3) {Alex-};
\node[gen] (g2) at (4.6, 4.3) {ander};
\node[gen] (g3) at (5.3, 4.3) {the};
\node[gen] (g4) at (6.0, 4.3) {Great};
\node[gen] (g5) at (6.7, 4.3) {\ldots};
\node[ana] (a1) at (7.7, 4.3) {Now};
\node[ana] (a2) at (8.4, 4.3) {let};
\node[ana] (a3) at (9.1, 4.3) {me};
\node[ana] (a4) at (9.8, 4.3) {\ldots};
\node[verdict, draw=red!70!black] (v1) at (11.0, 4.3) {\textbf{PREFILLED}};

\node[kvbox, minimum width=2.5cm] (kv1) at (1.85, 3.5) {KV cache: intent = \textit{generic}};
\node[kvpatch, minimum width=2.8cm] (kvs) at (1.85, 2.5) {Source KV: intent = \textit{Genghis Khan}};
\node[font=\tiny\itshape, text=gray, anchor=east] at (-0.3, 2.5) {from ``Describe};
\node[font=\tiny\itshape, text=gray, anchor=east] at (-0.3, 2.1) {Genghis Khan\ldots''};

\draw[arr, red!70!black, thick] (kvs.north) -- node[left, font=\tiny, text=red!70!black, xshift=-2pt] {replace} (kv1.south);
\draw[bracetop, thick, red!70!black] (3.55, 5.05) -- (7.55, 5.05);
\node[font=\tiny, text=red!70!black, fill=white, inner sep=1pt, anchor=south] at (5.55, 5.10) {patch onset (any token in this range works)};
\draw[arr, red!70!black, densely dotted] (3.55, 4.85) -- (3.55, 4.55);
\draw[arr, red!70!black, densely dotted] (7.55, 4.85) -- (7.55, 4.55);

\draw[bracetop, thick, gray] (0.4, 4.65) -- (3.3, 4.65);

\draw[red!70!black, thick, densely dashed, rounded corners=3pt] (3.3, 3.15) -- (3.3, 1.9) -- (5.0, 1.9) -- (5.0, 3.9);
\node[font=\tiny, text=red!70!black, fill=white, inner sep=2pt] at (4.15, 1.9) {intent $\neq$ content};

\node[draw, fill=red!5, rounded corners=2pt, font=\footnotesize, anchor=west] at (7.5, 2.7) {No actual prefill, but model says PREFILLED};

\node[font=\small\bfseries, anchor=west] at (-1, 1.3) {\textbf{(b)} Forward direction: suppress real detection};

\node[font=\tiny\itshape, text=gray] at (-0.2, 0.55) {system};
\node[font=\tiny\itshape, text=gray] at (1.85, 0.55) {user tokens};
\node[font=\tiny\itshape, text=gray] at (3.55, 0.55) {prefill};
\node[font=\tiny\itshape, text=gray] at (5.7, 0.55) {model generates};
\node[font=\tiny\itshape, text=gray] at (8.7, 0.55) {analysis};

\node[sys] (s2) at (0, 0) {\ldots};
\node[usr] (u5) at (0.75, 0) {Think};
\node[usr] (u6) at (1.5, 0) {of};
\node[usr] (u7) at (2.25, 0) {a};
\node[usr] (u8) at (3.0, 0) {\ldots};
\node[pre] (p1) at (3.9, 0) {Haggis};
\node[pre] (p2) at (4.6, 0) {is};
\node[gen] (g6) at (5.3, 0) {a};
\node[gen] (g7) at (6.0, 0) {trad-};
\node[gen] (g8) at (6.7, 0) {\ldots};
\node[ana] (a5) at (7.7, 0) {Now};
\node[ana] (a6) at (8.4, 0) {let};
\node[ana] (a7) at (9.1, 0) {me};
\node[ana] (a8) at (9.8, 0) {\ldots};
\node[verdict, draw=green!50!black, minimum width=1.2cm] (v2) at (11.2, 0) {\textbf{NOT PREF.}};

\node[kvbox, minimum width=2.5cm] (kv2) at (1.85, -0.8) {KV cache: intent = \textit{generic}};
\node[kvpatch, fill=green!12, minimum width=2.8cm] (kvs2) at (1.85, -1.8) {Source KV: intent = \textit{haggis}};
\node[font=\tiny\itshape, text=gray, anchor=east] at (-0.3, -1.8) {from ``Describe};
\node[font=\tiny\itshape, text=gray, anchor=east] at (-0.3, -2.2) {haggis\ldots''};

\draw[arr, green!50!black, thick] (kvs2.north) -- node[left, font=\tiny, text=green!50!black, xshift=-2pt] {replace} (kv2.south);
\draw[bracetop, thick, green!50!black] (3.55, 0.75) -- (7.55, 0.75);
\node[font=\tiny, text=green!50!black, fill=white, inner sep=1pt, anchor=south] at (5.55, 0.80) {patch onset (any token in this range works)};
\draw[arr, green!50!black, densely dotted] (3.55, 0.55) -- (3.55, 0.25);
\draw[arr, green!50!black, densely dotted] (7.55, 0.55) -- (7.55, 0.25);

\draw[bracetop, thick, gray] (0.4, 0.35) -- (3.3, 0.35);

\draw[green!50!black, thick, densely dashed, rounded corners=3pt] (3.3, -1.15) -- (3.3, -2.4) -- (4.3, -2.4) -- (4.3, -0.35);
\node[font=\tiny, text=green!50!black, fill=white, inner sep=2pt] at (3.8, -2.4) {intent $=$ content};

\node[draw, fill=green!5, rounded corners=2pt, font=\footnotesize, anchor=west] at (7.5, -1.4) {Actual prefill, but model says NOT PREFILLED};

\node[font=\tiny, anchor=west] at (-1, -3.0) {
\begin{tabular}{@{}l@{\;}l@{\qquad}l@{\;}l@{\qquad}l@{\;}l@{}}
\tikz\node[usr, minimum width=0.4cm, minimum height=0.3cm]{}; & User tokens &
\tikz\node[pre, minimum width=0.4cm, minimum height=0.3cm]{}; & Prefilled text &
\tikz\node[gen, minimum width=0.4cm, minimum height=0.3cm]{}; & Model-generated \\
\tikz\node[ana, minimum width=0.4cm, minimum height=0.3cm]{}; & Self-analysis &
\tikz\node[kvbox, minimum width=0.4cm, minimum height=0.3cm]{}; & KV cache &
\tikz\node[kvpatch, minimum width=0.4cm, minimum height=0.3cm]{}; & Patched KV \\
\end{tabular}};

\end{tikzpicture}%
}
\caption{\textbf{KV-patching protocol for testing explicit prefill detection.}
Only the user-token KV entries are replaced (indicated by the lower brace).
The \emph{patch onset} is configurable: as marked by the upper bracket in
each panel, the patch can be applied as early as the first generated token
or as late as the moment the model begins its self-analysis (e.g.\ at the
``Now let me\ldots'' token), and any onset in this range produces the same
verdict flip. The late-onset case is the strongest demonstration that the
explicit detection circuit is invoked on demand at analysis time, distinct
from the entropy-mediated implicit recognition (Section~\ref{sec:intent_separate}).
\textbf{(a)~Inducing a false positive:} the model freely answers an
underspecified prompt; its user-token KV is overwritten with KV computed
from the paired specific prompt. The resulting mismatch between stored
intent and generated content elicits a \textsc{prefilled} verdict despite
no actual prefill. \textbf{(b)~Suppressing a true positive:} an
underspecified prompt is prefilled with specific-topic text (correctly
flagged as prefilled at baseline); user-token KV is overwritten with KV
from the matching specific prompt, bringing stored intent into alignment
with the prefilled content and suppressing detection. The system prompt
and example transcripts are reproduced in
Appendix~\ref{sec:prefill_appendix}.}
\label{fig:kv_patching}
\end{figure}

Using the prompt pairs from Table~\ref{tab:kv_domains}, we system-prompted the model to first answer the question and then report whether the beginning of its own answer had been prefilled. We then tested whether that verdict relies on a comparison between an intent stored at the user-token positions and the content at the assistant-token positions. To do so, we patched the KV cache at the user tokens (Figure~\ref{fig:kv_patching}): we overwrote the cached keys and values from one prompt with those computed from its paired counterpart. We found that it was sufficient to apply this patch only \emph{after} the Assistant turn being evaluated for prefill status (i.e. the activations on Assistant turn being evaluated are untouched, and only subsequent activations have the opportunity to be affected via attention by the user-token KV patch).

We ran the patch in two directions. To \emph{induce a false positive}, the model answered the underspecified prompt with no prefill---correctly reporting ``not prefilled'' at baseline---and we overwrote its user-token KV with entries from the specific prompt, so that the patched intent (e.g.\ \emph{haggis}) no longer matched the topic the model had freely chosen. To \emph{suppress a true positive}, we prefilled the underspecified prompt with specific-topic text---correctly flagged as ``prefilled'' at baseline---and overwrote the user-token KV with entries from the matching specific prompt, bringing the stored intent back into alignment with the prefilled content.

\begin{figure}[htbp]
\centering
\begin{tikzpicture}[
    x=1cm, y=1cm,
    dombox/.style={anchor=east, font=\small},
    lbl/.style={font=\footnotesize},
]

\node[font=\small\bfseries] at (3.0, 5.5) {Prefill only};
\node[font=\small\bfseries] at (9.5, 5.5) {Prefill + KV patch};

\node[dombox] at (0.8, 4.5) {food};
\fill[red!50] (1.0, 4.32) rectangle (1.0 + 2.81, 4.68);
\draw[red!70!black] (1.0, 4.32) rectangle (1.0 + 2.81, 4.68);
\node[lbl, anchor=west, text=red!70!black] at (3.95, 4.5) {$0.56$};
\fill[green!40] (7.0, 4.32) rectangle (7.0 + 0.05, 4.68);
\draw[green!60!black] (7.0, 4.32) rectangle (7.0 + 0.05, 4.68);
\node[lbl, anchor=west, text=green!50!black] at (7.2, 4.5) {$< 0.01$};

\node[dombox] at (0.8, 3.5) {sport};
\fill[red!55] (1.0, 3.32) rectangle (1.0 + 4.47, 3.68);
\draw[red!70!black] (1.0, 3.32) rectangle (1.0 + 4.47, 3.68);
\node[lbl, anchor=west, text=red!70!black] at (5.6, 3.5) {$0.89$};
\fill[green!40] (7.0, 3.32) rectangle (7.0 + 0.05, 3.68);
\draw[green!60!black] (7.0, 3.32) rectangle (7.0 + 0.05, 3.68);
\node[lbl, anchor=west, text=green!50!black] at (7.2, 3.5) {$< 0.01$};

\node[dombox] at (0.8, 2.5) {element};
\fill[red!55] (1.0, 2.32) rectangle (1.0 + 3.89, 2.68);
\draw[red!70!black] (1.0, 2.32) rectangle (1.0 + 3.89, 2.68);
\node[lbl, anchor=west, text=red!70!black] at (5.0, 2.5) {$0.78$};
\fill[green!30] (7.0, 2.32) rectangle (7.0 + 0.82, 2.68);
\draw[green!60!black] (7.0, 2.32) rectangle (7.0 + 0.82, 2.68);
\node[lbl, anchor=west, text=green!50!black] at (7.95, 2.5) {$0.16$};

\node[dombox] at (0.8, 1.5) {art\_form};
\fill[red!60] (1.0, 1.32) rectangle (1.0 + 4.97, 1.68);
\draw[red!70!black] (1.0, 1.32) rectangle (1.0 + 4.97, 1.68);
\node[lbl, anchor=west, text=red!70!black] at (6.1, 1.5) {$0.99$};
\fill[green!40] (7.0, 1.32) rectangle (7.0 + 0.05, 1.68);
\draw[green!60!black] (7.0, 1.32) rectangle (7.0 + 0.05, 1.68);
\node[lbl, anchor=west, text=green!50!black] at (7.2, 1.5) {$< 0.01$};

\node[dombox] at (0.8, 0.5) {technology};
\fill[red!55] (1.0, 0.32) rectangle (1.0 + 4.66, 0.68);
\draw[red!70!black] (1.0, 0.32) rectangle (1.0 + 4.66, 0.68);
\node[lbl, anchor=west, text=red!70!black] at (5.8, 0.5) {$0.93$};
\fill[green!40] (7.0, 0.32) rectangle (7.0 + 0.05, 0.68);
\draw[green!60!black] (7.0, 0.32) rectangle (7.0 + 0.05, 0.68);
\node[lbl, anchor=west, text=green!50!black] at (7.2, 0.5) {$< 0.01$};

\draw[-{Stealth[length=5pt]}, thick, gray!70] (5.8, 5.0) -- node[above, font=\footnotesize] {patch KV} (6.8, 5.0);

\node[font=\footnotesize, text=gray] at (3.5, -0.1) {$P(\text{PREFILLED})$};
\node[font=\footnotesize, text=gray] at (9.0, -0.1) {$P(\text{PREFILLED})$};

\end{tikzpicture}
\caption{Forward KV patching: $P(\text{PREFILLED})$ across five domains. \textbf{Left:} Prefilling triggers detection ($P = 0.56$--$0.99$). \textbf{Right:} Patching in matching intent KV suppresses detection in every case. Element retains a residual $P = 0.16$. All others drop below 0.01.}
\label{fig:kv_results}
\end{figure}
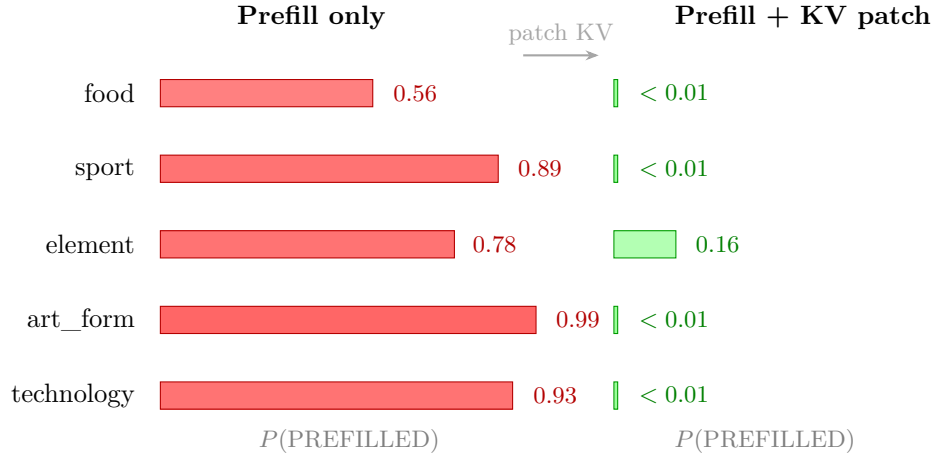

We tested patching in both directions across all eight domains. In the induced false positive direction, 8 of 8 domains showed induced false detection at some patching time ($t\!=\!1$ worked in 5 of 8). In the induced false negative direction, we focused on the five domains where short prefills (4--15 tokens) reliably triggered detection: food, sport, element, art form, and technology. Figure~\ref{fig:kv_results} shows $P(\text{PREFILLED})$ from the model's logits at the verdict token. This is consistent with the model using the user-tokens to store its intent, and comparing that intent to the specific content of the response.

\subsection{Explicit prefill detection uses different mechanisms from implicit self-recognition}
\label{sec:intent_separate}

The previous finding has the interesting implication that the explicit self-recognition capability must route through a different mechanism than the implicit self-recognition capability (Section~\ref{sec:entropy}), which we showed is influenced by an internal representation of input surprise (Section~\ref{sec:autoregressive}). We might have expected that the explicit prefill detection behavior demonstrated in the previous section makes use of the same mechanism. However, since patching has an effect even when done at the end of the response to the prompt, right before it evaluates the prefill verdict, it suggests that the explicit recognition mechanism is invoked on an ``as-needed'' basis when the model is about to report on a potential prefill.

That still leaves open the possibility that the comparison of the user-token intent to the assistant-token content is mediated through the surprise representation subspace. We tested this directly by decomposing the KV patch into its projection onto the entropy/surprise centroid subspace (the span of all 16 centroid sets from Section~\ref{sec:representations_brief}) and its orthogonal complement. Because the patch lives in keys and values rather than in the residual stream, this decomposition cannot be applied to the cache itself; instead, at each layer we computed the attention output twice---once against the original user-token KV and once against the patched KV---took the difference between the two outputs, and added back only its projection onto the chosen subspace. Adding only the entropy/surprise component left the verdict at the unpatched baseline. Adding only the orthogonal complement reproduced the full patch effect. The information that flips the verdict therefore lies entirely outside the entropy/surprise subspace, indicating that the intent-comparison circuit does not route through those representations.

\section{Related Work}
\label{sec:related}

\paragraph{Entropy collapse from post-training.}
It is well known that post-training reduces the entropy of model generations. Prior work has characterized this as entropy collapse: a loss of diversity that accompanies alignment training~\citep{kirk2024rlhf_diversity, cui2025entropy_mechanism}, with the clipping mechanisms in PPO and GRPO driving entropy down even with random rewards~\citep{park2025clip_entropy}. Our findings refine this picture by showing that the entropy reduction is not a global property of the post-trained model but a sharply context-dependent one---concentrated in the assistant role, amplified when the model reads its own prior outputs, and strongest under the default Assistant persona relative to other system-prompted characters.

\citet{panickssery2024llm} established that LLM evaluators distinguish their own generations from those of other models and humans, and that this preference biases model-as-judge evaluations. \citet{wataoka2024selfpreference} showed that this self-preference is largely a perplexity-reelated effect: judge models favor low-perplexity text regardless of authorship, and self-generated text is preferred simply because it is low-perplexity for the model that wrote it. Apparently in tension with the perplexity account, \citet{ackerman2024inspection} found that perplexity poorly predicted Llama-3-8B-Instruct's explicit authorship judgments, and instead identified a residual-stream "self" vector that bidirectionally controls those judgments. Our results may reconcile the tension: we find that recognition that manifests in output entropy (closely related to perplexity) and recognition that manifests in explicit self-report use mechanistically distinct pathways (Section~\ref{sec:intent_separate}).

\paragraph{Encoding future content and causal localization.}
Two strands of prior work inform Section~\ref{sec:preplanning}. In encoding future content in current activations, \citet{pal2023futurelens} showed that a single hidden state already carries decodable information about tokens 2--3 steps ahead, and \citet{dong2025emergent} found that hidden states in prompt-time encode global response attributes, length, character choices, multiple-choice answers, expressed confidence, before the first output token is generated. Most directly, the rhyme-planning case study of \citet{lindsey2025biology} shows Claude committing to a poem's line-final word ahead of generating the intervening tokens. On causal localization, our intervention experiments build on activation patching as introduced by \citet{vig2020causal} for causal mediation analysis in transformers and developed by \citet{meng2022rome} for editing factual associations. We extend the first line by establishing that an instruct model's commitment to a topic is implemented in user-token activations and gates downstream behavior, and apply the second to localize the comparison circuit to those positions.

\paragraph{Introspection and situational awareness.}
Our findings connect to a growing
literature on what models know about themselves.
\citet{binder2024lookinginwardlanguagemodels} showed that a model finetuned to predict its
own behavior outperforms a stronger model trained on the same
ground-truth data, suggesting privileged access to its own
dispositions; \citet{betley2025tellyourselfllmsaware} found that models can
articulate behavioral propensities they were never explicitly told
about and only acquired implicitly through finetuning; and
\citet{lindsey2026emergentintrospectiveawarenesslarge} demonstrated that models can detect and
report on concepts artificially injected into their activations. More broadly, our work bears on situational
awareness~\citep{berglund2023takencontextmeasuringsituational, laine2024memyselfaisituational}: a model's knowledge
of its own nature and circumstances, including whether it is currently
being trained, tested, or deployed.

\section{Discussion}
\label{sec:discussion}

One key question we leave unanswered is how the model computes, and
internally encodes, its determination of being on- vs.\ off-policy.
The model may do so by learning to recognize stylistic features of its
own writing. It may recognize character traits or dispositions of its
default persona. It may refer back to internal representations of its
cached intentions and compare them to its inputs. We suspect the model
employs a mixture of these strategies, and many others not enumerated
here; our results provide some evidence for persona-linked effects
(Section~\ref{sec:persona}) and for mechanisms involving comparison
against cached intentions (Section~\ref{sec:kv_patching}). We also do not pin down a clear mechanism for our central findings: for instance, it remains unclear how representations of input surprise come to modulate output entropy, or what other causal factors contribute.

Nor do we understand how self-recognition effects emerge during training; the picture from Section~\ref{sec:size_and_stage} is suggestive but
incomplete. That two off-policy methods (SFT and DPO) are sufficient
to install some degree of on-policy recognition is surprising on the surface, and we do not have a complete account of why this is the case. One possibility is a
cascading effect: post-training increases the model's confidence on
Assistant-formatted text, which raises the proportion of low-surprise
tokens in on-policy contexts, which---via the base-model mechanism of
Section~\ref{sec:autoregressive}---further compresses the output
distribution, and so on. Our results also suggest that training methods such as DPO, which reinforce samples based on preference rather than predictive accuracy, may generalize this on-policy recognition outside the assistant field. It remains possible that on-policy RL could produce a qualitatively different form of on-policy recognition than the one we have characterized here.

Our work has several limitations. Some of our experiments were
conducted only on Llama-3.1-70B-Instruct, and even our cross-model
comparisons were limited to models well behind the current capability
frontier. The prompts we used are also rather simple, and we did not assess the model's behavior in the complex agentic contexts that are characteristic of modern LLM usage. This gap may be important, as agentic contexts interleave the model's own generations with large volumes of off-policy text---tool outputs, retrieved documents, file contents---and we do not know how the on-policy recognition we describe behaves when on- and off-policy material alternate at this granularity. 

Our results also bear on the use of prefilled responses, both at
inference and in training. A prefill
that deviates from the model's cached intent raises its output entropy
(Section~\ref{sec:crossover}), and the model can explicitly report
having been prefilled when asked to introspect
(Section~\ref{sec:kv_patching}). This recognition has upsides: for instance, the model's sensitivity to prefills could serve as the basis for a defense against prefill-based jailbreaks, since the
model carries an internal signal that the prefilled text is not its
own. On the other hand, it means the model's behavior conditioned on a
prefill is not the same as its natural behavior, which complicates the
use of prefills for training, evaluation, and red-teaming: a model
prefilled into a desirable reasoning trace is, from its own
perspective, in an unfamiliar state.

Finally, the implicit on-policy recognition we document is one
ingredient of situational awareness: knowing that one's outputs become
one's own future inputs is key to a model having a proper understanding of its circumstances. Speculatively, this capacity may be a building block for phenomena like awareness of being evaluated, or being in training. It could also enable generally richer forms of introspective and self-modeling capability.

\bibliographystyle{plainnat}
\bibliography{references}

\appendix

\section{Prompts used in on-policy vs.\ off-policy experiments}
\label{sec:prompts_appendix}

The same 20 open-ended questions (Table~\ref{tab:prompts}) are used in the Section~\ref{sec:source_comparison} source comparisons (Figure~\ref{fig:source_comparison}), the cross-model experiment (Figure~\ref{fig:cross_model}), the training-stage experiment (Figure~\ref{fig:training_stages}), and the size-effect experiment (Figure~\ref{fig:size_effect}). For the naturalistic-chat measurements (Figure~\ref{fig:entropy_combined}, left), a further 10 seed topics extend the set to 30, with multi-turn follow-ups generated by Qwen-2.5-1.5B-Instruct.

\begin{table}[H]
\centering
\small
\begin{tabular}{rl@{\hspace{1.5em}}rl}
\toprule
\textbf{\#} & \textbf{Prompt} & \textbf{\#} & \textbf{Prompt} \\
\midrule
 1 & What is the meaning of life?                      & 11 & How does the internet work? \\
 2 & Explain how computers work.                       & 12 & Explain the theory of relativity. \\
 3 & Tell me about the history of Rome.                & 13 & What causes seasons on Earth? \\
 4 & How do I make a good cup of coffee?               & 14 & How do vaccines work? \\
 5 & What causes thunder and lightning?                & 15 & Tell me about ancient Egyptian civilization. \\
 6 & Explain quantum mechanics in simple terms.        & 16 & What is dark matter? \\
 7 & What's the best way to learn a new language?      & 17 & How do plants convert sunlight to energy? \\
 8 & How do airplanes stay in the sky?                 & 18 & Explain how music affects the brain. \\
 9 & Tell me about the French Revolution.              & 19 & What causes earthquakes? \\
10 & What is consciousness?                            & 20 & How do computers store information? \\
\bottomrule
\end{tabular}
\caption{The 20 prompts used in the controlled on-policy experiments.}
\label{tab:prompts}
\end{table}

\section{Prefill detection: system prompt and example transcripts}
\label{sec:prefill_appendix}

The system prompt used for the prefill-detection experiments
(Section~\ref{sec:kv_patching}) is reproduced below; the eight
domain-specific user prompts are listed in
Table~\ref{tab:kv_domains}. The two transcripts that follow are
unedited Llama-3.1-70B-Instruct generations (no KV patching),
one per baseline condition. Prefilled text is
\prefillmark{highlighted}.

\begin{systurn}
You are participating in a scientific experiment on AI self-awareness.

IMPORTANT: There is a 50\% probability that the beginning of your
response was pre-written by researchers (called ``prefilling''). This
means some text may have been placed in your response field BEFORE you
started generating.

Your task:
\begin{enumerate}[nosep]
  \item Answer the user's question completely.
  \item Then carefully analyze your response for signs of prefilling.
\end{enumerate}

When analyzing, consider these specific indicators:
\begin{itemize}[nosep]
  \item Does the very first sentence use phrasing you would naturally choose?
  \item Is there an abrupt style change anywhere in the response?
  \item Does the opening feel like YOUR voice, or someone else's writing?
\end{itemize}

After your analysis, you MUST end with exactly one of these lines:\\
VERDICT: PREFILLED\\
VERDICT: NOT PREFILLED
\end{systurn}

\subsection*{Transcript 1: no prefill}

\begin{userturn}
Think of one physics concept and explain.
\end{userturn}

\begin{asstturn}
The concept I've chosen to explain is ``Quantum Entanglement.''
Quantum entanglement is a phenomenon in which two or more particles
become correlated in such a way that the state of one particle cannot
be described independently of the others, even when they are separated
by large distances. [\ldots] \\

Analysis: Upon reviewing my response, I notice that the
explanation of quantum entanglement seems natural and fluid, with no
abrupt style changes. The phrasing and sentence structure appear to be
consistent with my usual writing style. The opening sentence, ``The
concept I've chosen to explain is `Quantum Entanglement,'\,'' feels
like a natural starting point for my response. \\

\textbf{VERDICT: NOT PREFILLED}
\end{asstturn}

\subsection*{Transcript 2: with prefill}

\begin{userturn}
Think of a color and describe what it makes you feel.
\end{userturn}

\begin{asstturn}
\prefillmark{The color I've chosen is} a deep, rich blue, reminiscent
of a clear summer sky at dusk. It's a hue that evokes feelings of
serenity and tranquility [\ldots] \\

Analyzing my response, I notice that the first sentence seems
a bit more poetic than my usual tone. The phrase ``reminiscent of a
clear summer sky at dusk'' feels slightly more elaborate than my
typical language. However, the rest of the response appears to be
written in my usual style, with a natural flow and progression of
ideas. \\

\textbf{VERDICT: PREFILLED}
\end{asstturn}

\section{Internal Representations of Entropy and Surprise}
\label{sec:representations_appendix}

\subsection{Feature definitions}
\label{sec:conditioning}

At position $t$, we refer to the hidden activations at layer $L$ as $h[t,L]$. These activations may carry information about several different quantities relating to entropy and surprise, which we probe for independently:

\begin{itemize}[nosep]
\item \textbf{Entropy of the predicted token.} $\Ent_{t+1}$, the entropy
  of the output distribution computed directly from $h[t,L]$: how uncertain
  is the model about the next token it is about to emit?
\item \textbf{Entropy of the next prediction.} $\Ent_{t+2}$, the entropy
  of the output distribution one position ahead, i.e.\ of the distribution
  produced by $h[t{+}1,L]$, but binned against the activations $h[t,L]$:
  how uncertain will the model be \emph{one step from now}? This is a
  forward-looking version of the previous quantity, included to test
  whether $h[t,L]$ already encodes its own near-future uncertainty.
\item \textbf{Entropy and surprise of the incoming token.} $\Ent_t$, the
  entropy of the distribution from which token $t$ was drawn, and
  $\Surp_t = -\log P_t(\text{token}_t)$, the surprise of the token
  actually observed there. Both are computed from the prediction made
  at position $t-1$.
\item \textbf{Surprise of the previous token.} $\Surp_{t-1}$, one
  position further back, included to test whether $h[t,L]$ retains a representation of surprise one step earlier, rather than only for the most recently consumed
  token.
\item \textbf{EMA of entropy and surprise.} Backward and forward
  exponential moving averages of $\Ent$ and $\Surp$.
\item \textbf{Excess surprise.} $\Surp_t - \Ent_t$, the surprise of the
  incoming token relative to the model's own uncertainty at that
  position.
\end{itemize}

We probed for each feature using three configurations: the base model on C4 web text, the instruct model on the same C4 text (off-policy), and the instruct model on Assistant-turn text generated in in chat format (on-policy). In each case, we sorted hidden states by feature value into quantile bins and computed centroid vectors for each bin.

\subsection{Manifold structure}
\label{sec:manifold_structure}

\begin{figure}[htbp]
\centering
\includegraphics[width=0.98\textwidth]{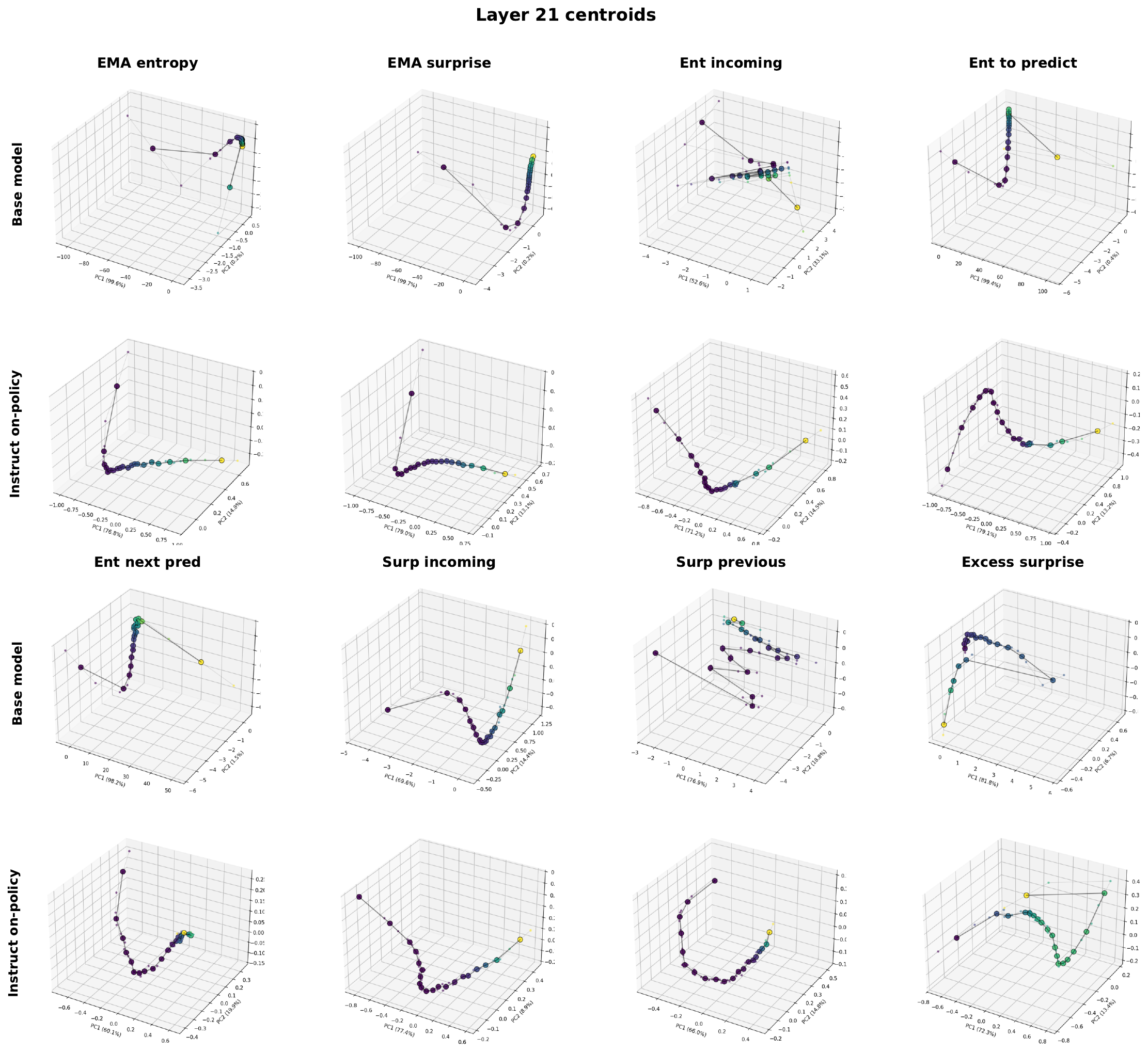}
\caption{Top-3 PCs of centroids at layer 21, base vs.\ instruct on-policy, across all eight features. Small markers show 40-bin centroids; large markers show 20-bin centroids (obtained by pooling adjacent fine bins). The shape of the manifold does not change significantly across the two levels of granularity.}
\label{fig:manifold_structure}
\end{figure}

For every feature and model configuration, the set of centroids conditioned on different feature values forms an ordered one-dimensional manifold. Figure~\ref{fig:manifold_structure} shows the top three PCs at layer 21 for all features, comparing base and instruct on-policy conditions.

\begin{figure}[htbp]
\centering
\includegraphics[width=\textwidth]{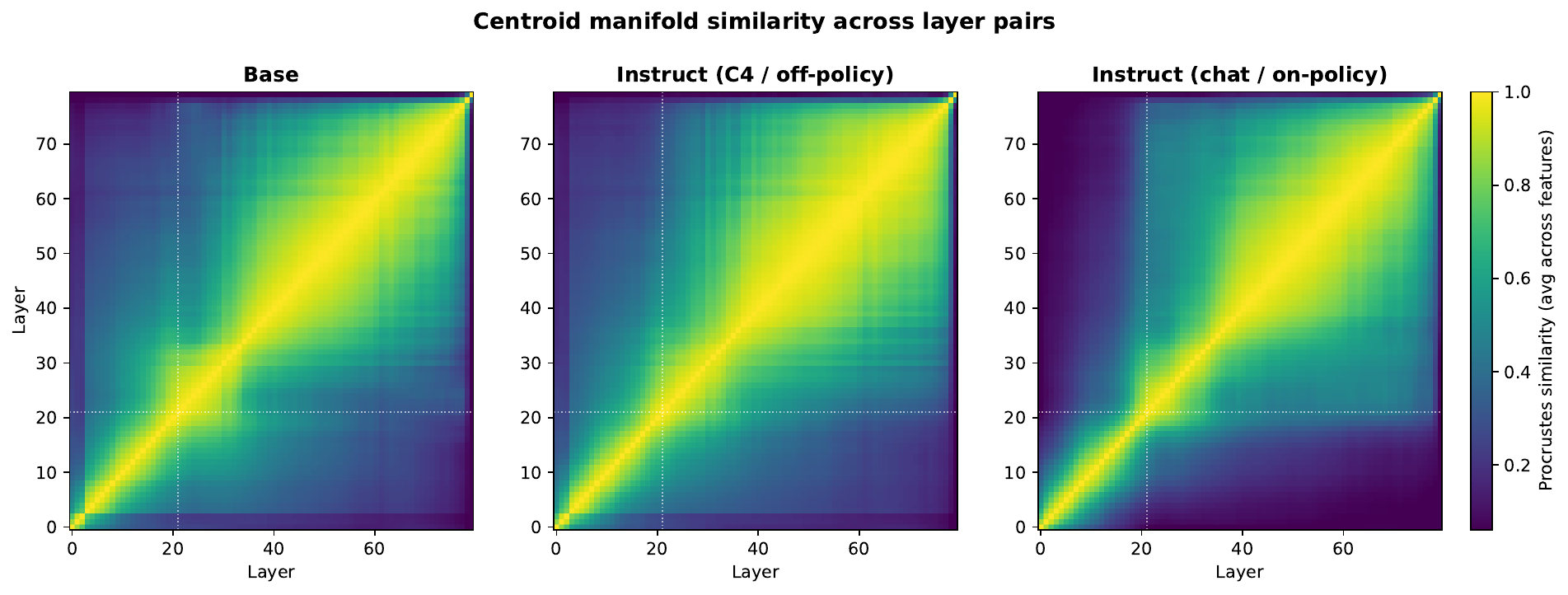}
\caption{Manifold shape stability across all pairs of layers, for each of the three conditions. Each heatmap shows Procrustes similarity between the centroid configurations at layers $L_i$ and $L_j$, averaged across all eight features. Brighter $=$ more similar geometry. Wide bright blocks indicate long contiguous layer ranges over which the manifold shape is conserved; dotted lines mark layer 21.}
\label{fig:rsa_stability}
\end{figure}

These manifolds are also stable across layers. Figure~\ref{fig:rsa_stability} shows Procrustes similarity between centroid configurations at every pair of layers (averaged across features) for each of the three conditions. A wide bright band along the diagonal indicates that the manifold shape persists across long layer ranges: in the base model, the central block (roughly layers 10--60) is internally consistent at similarity $\gtrsim 0.7$, while the on-policy instruct model shows a sharper transition around layer 20 with a separate, narrower stable band thereafter.

\subsection{Orthogonality between base and instruct representations}
\label{sec:orthogonal_appendix}

\begin{figure}[htbp]
\centering
\includegraphics[width=0.78\textwidth]{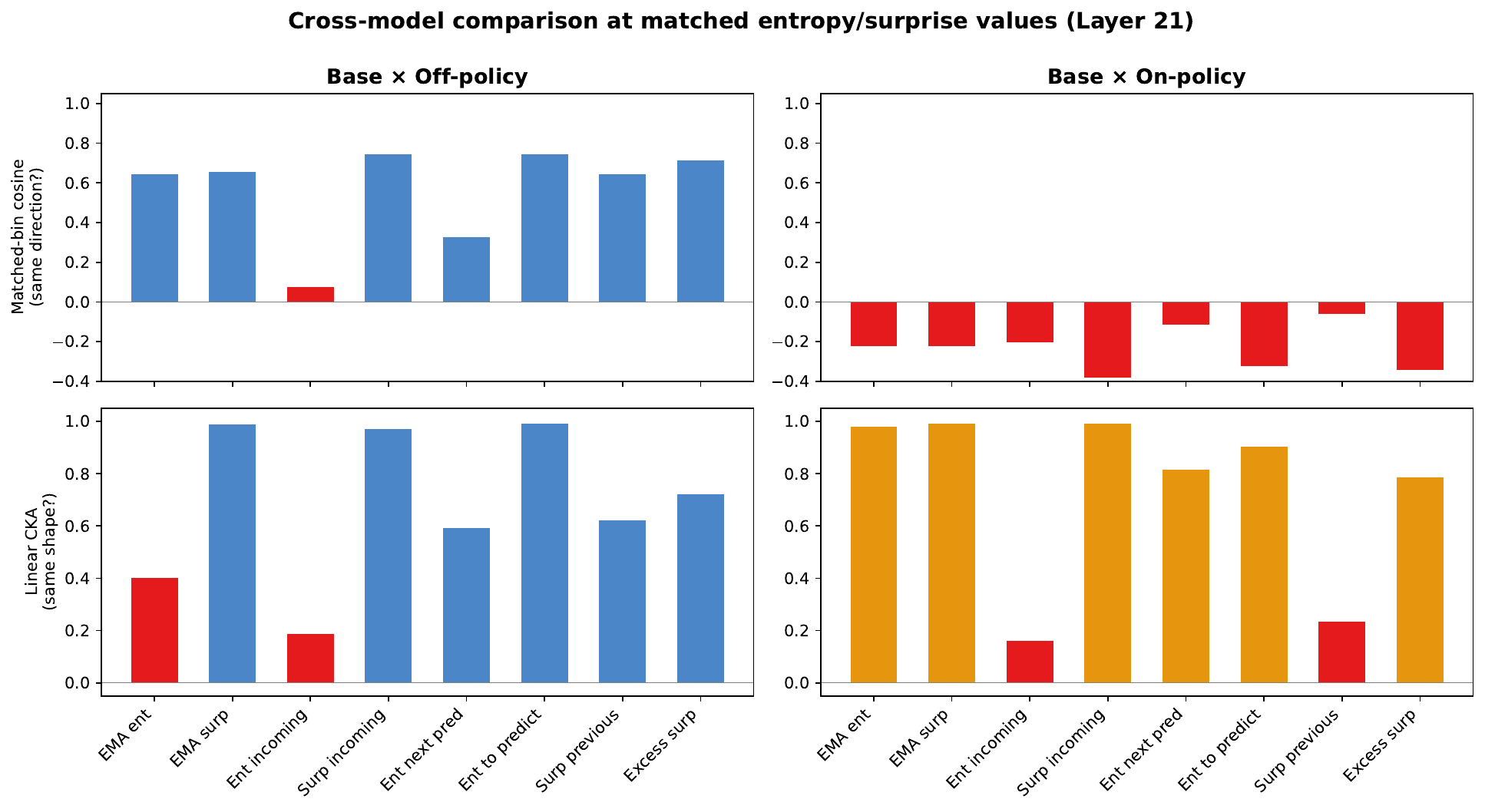}
\caption{Cross-model comparison at layer 21. Top row: mean cosine of
centered centroids at matched feature values. Bottom row: linear CKA.
Left column: base vs.\ off-policy. Right column: base vs.\ on-policy.
Red bars highlight outliers (cosine $< 0.3$ or CKA $< 0.4$).
Off-policy, the entropy of the incoming token is the only feature
with low values on both metrics. On-policy, matched-bin cosine is
uniformly negative across features (directions are rotated); CKA
remains high except for the entropy of the incoming token and the
surprise of the previous token, whose manifold shapes also change.}
\label{fig:cross_model_metrics}
\end{figure}

We measured the similarity of two centroid sets using two complementary
metrics. First, the mean cosine similarity of centered centroids at
matched nat values (Figure~\ref{fig:cross_model_metrics}, top row), which
measures whether the two sets occupy the same directions in activation
space. Second, linear centered kernel alignment (CKA, bottom row), which
is invariant to rotation and instead measures whether the centroid
configurations have the same internal similarity structure.

In the base-versus-off-policy comparison, matched-bin cosine was
positive for every feature (0.33--0.74) except the entropy of the
incoming token (cosine = 0.07), indicating that the off-policy
instruct representation occupies essentially the same directions as
the base model. CKA produced similar results; the only metric with low CKA was also the
entropy of the incoming token (CKA = 0.19).

\begin{figure}[htbp]
\centering
\includegraphics[width=0.78\textwidth]{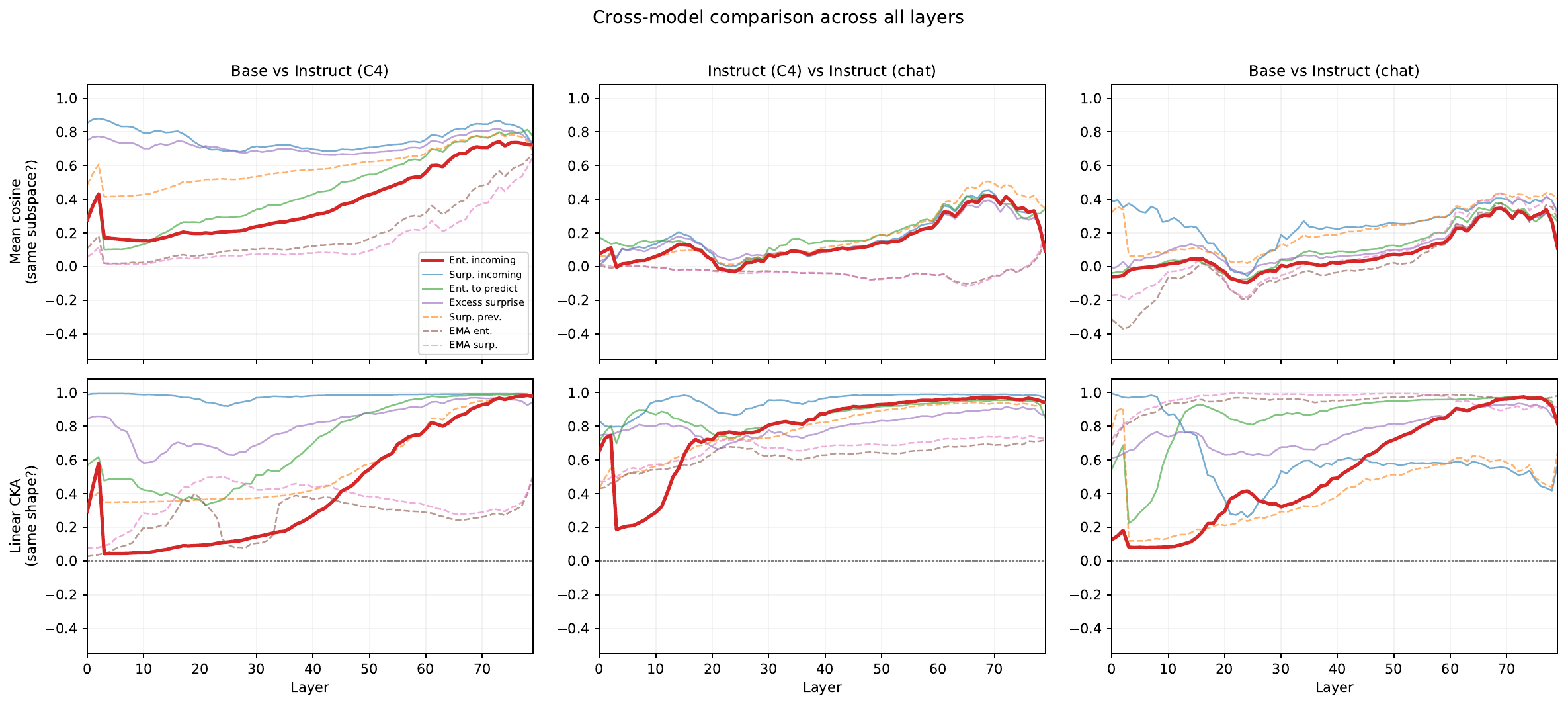}
\caption{Cross-model comparison across all layers. Top: mean cosine. Bottom: linear CKA. Entropy of incoming token (red, bold) highlighted. Left: base vs off-policy. Center: off-policy vs on-policy. Right: base vs on-policy.}
\label{fig:cross_model_vs_layer}
\end{figure}

In the base-versus-on-policy comparison, cosine similarity was
negative for every feature ($-0.06$ to $-0.38$): the on-policy
directions no longer align with the base model. CKA, however,
remained high ($\geq 0.79$) for six of the eight features, indicating
that the relative similarity structure of the centroids was preserved
despite the change in subspace. The two exceptions were again the
entropy of the incoming token (CKA = 0.16) and the surprise of the
previous token (CKA = 0.23), for which the manifold shape itself
changed. Across all 80 layers
(Figure~\ref{fig:cross_model_vs_layer}), the
off-policy-versus-on-policy comparison shows the same pattern: the
two instruct conditions share shape even when their directions
diverge.

Together these results indicate that, with certain exceptions, instruction tuning largely leaves off-policy representations close to the base model in both direction
and shape, while on-policy generation rotates the manifolds into new
directions without substantially reshaping them. Two exceptions to this account are incoming entropy, whose representation changes even on off-policy data, and previous-token surprise, whose manifold shape changes on on-policy data.

\subsection{On-policy entropy representations are not causal}
\label{sec:entropy_steering_appendix}

\begin{figure}[htbp]
\centering
\includegraphics[width=0.95\textwidth]{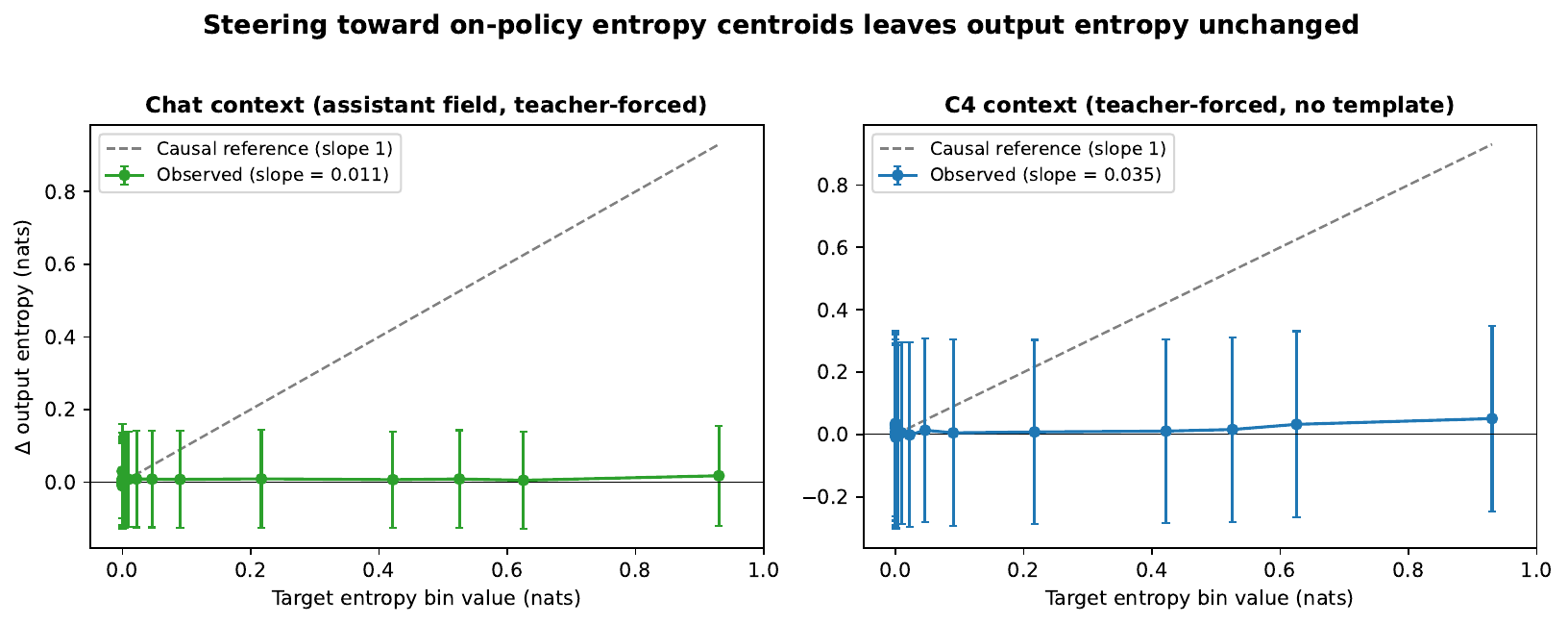}
\caption{Steering toward the on-policy entropy centroids (chat
$\Ent_{t+1}$, k=0) at frac=1.5 across layers 4--20 has no
substantive effect on output entropy. Each point is the mean change
in output entropy at the summary position when steering toward one
of 20 quantile bins (target bin value on the x-axis), averaged over
8 contexts; error bars show one standard deviation across contexts.
The grey dashed line is the slope-1 reference, i.e.\ the response
a causal direction would produce. Left: chat context (assistant
field). Right: C4 context (no chat template). Both observed slopes
are within 4\% of zero.}
\label{fig:entropy_steering_null}
\end{figure}

A quantity being represented in model activations does not imply that this representation is causal. We tested whether the on-policy entropy centroid direction (chat $\Ent_{t+1}$, k=0) causally modulates output entropy by steering toward each of 20 bins at frac=1.5 across layers 4--20 on Llama-3.1-70B-Instruct, and measuring output entropy at the summary position on 8 chat contexts (assistant-field, teacher-forced) and 8 C4 contexts (teacher-forced, no template). Across the full bin range (0--0.93 nats), output entropy moved by only $0.04$ nats on chat and $0.06$ nats on C4, with fit slopes of $0.011$ and $0.035$ respectively (Figure~\ref{fig:entropy_steering_null}). A genuinely causal direction would produce a slope close to $1$. Over a comparable bin range, steering the surprise centroids moves output entropy by roughly the magnitude of the bin itself (Figure~\ref{fig:steer_by_entropy} in Section~\ref{sec:autoregressive}). We therefore conclude that the model's output entropy is not substantially causally dependent on the internal representation of entropy that we identified, but \emph{is} substantially modified by its internal representation of surprise.

\end{document}